%% file: main.tex
\documentclass[10pt,a4paper]{article}
\input{preamble}

\title{Toward Real-Time VLAs: Stage-Aware Two-Step Flow Denoising and System-Level Evaluation}
\newcommand{\reportauthors}{%
    Di Wu\textsuperscript{1,3}\quad
    Rongtian Shen\textsuperscript{1}\quad
    Ping Liu\textsuperscript{1}\quad
    Yan Shen\textsuperscript{1,4}\quad
    Zhenhan Yin\textsuperscript{1,5}\quad
    Shun Zuo\textsuperscript{1,6}\\[2pt]
    Xuhua Chen\textsuperscript{1}\quad
    He Zheng\textsuperscript{1}\quad
    Lingfeng Zhang\textsuperscript{1}\quad
    Jiangning Zhang\textsuperscript{2}\quad
    Tao Zhang\textsuperscript{1,*}%
}
\author{\reportauthors}
\date{Technical Report}

\begin{document}
\vspace*{-18pt}
\begin{center}
{\setstretch{1.0}\magictitlefont\fontsize{16}{20}\selectfont\bfseries%
Toward Real-Time VLAs: Stage-Aware Two-Step\\
Flow Denoising and System-Level Evaluation\par}
\vspace{1.0em}
{\setstretch{1.0}\setlength{\parskip}{0pt}%
\fontsize{11}{13}\selectfont\reportauthors\par
\vspace{3pt}
\fontsize{9}{11}\selectfont
\textsuperscript{1}Magic-Lab Team, Magiclab Robotics Inc.\quad
\textsuperscript{2}Zhejiang University\quad
\textsuperscript{3}Southeast University\\
\textsuperscript{4}Harbin Institute of Technology\quad
\textsuperscript{5}Tongji University\quad
\textsuperscript{6}Jilin University\par}
\vspace{0.75em}
\end{center}
\begingroup
\renewcommand{\thefootnote}{\fnsymbol{footnote}}
\footnotetext[1]{Corresponding author and project lead.}
\endgroup
\thispagestyle{plain}
\input{sections/00_abstract_en}
\begin{figure}[H]
    \centering
    \includegraphics[width=0.90\linewidth]{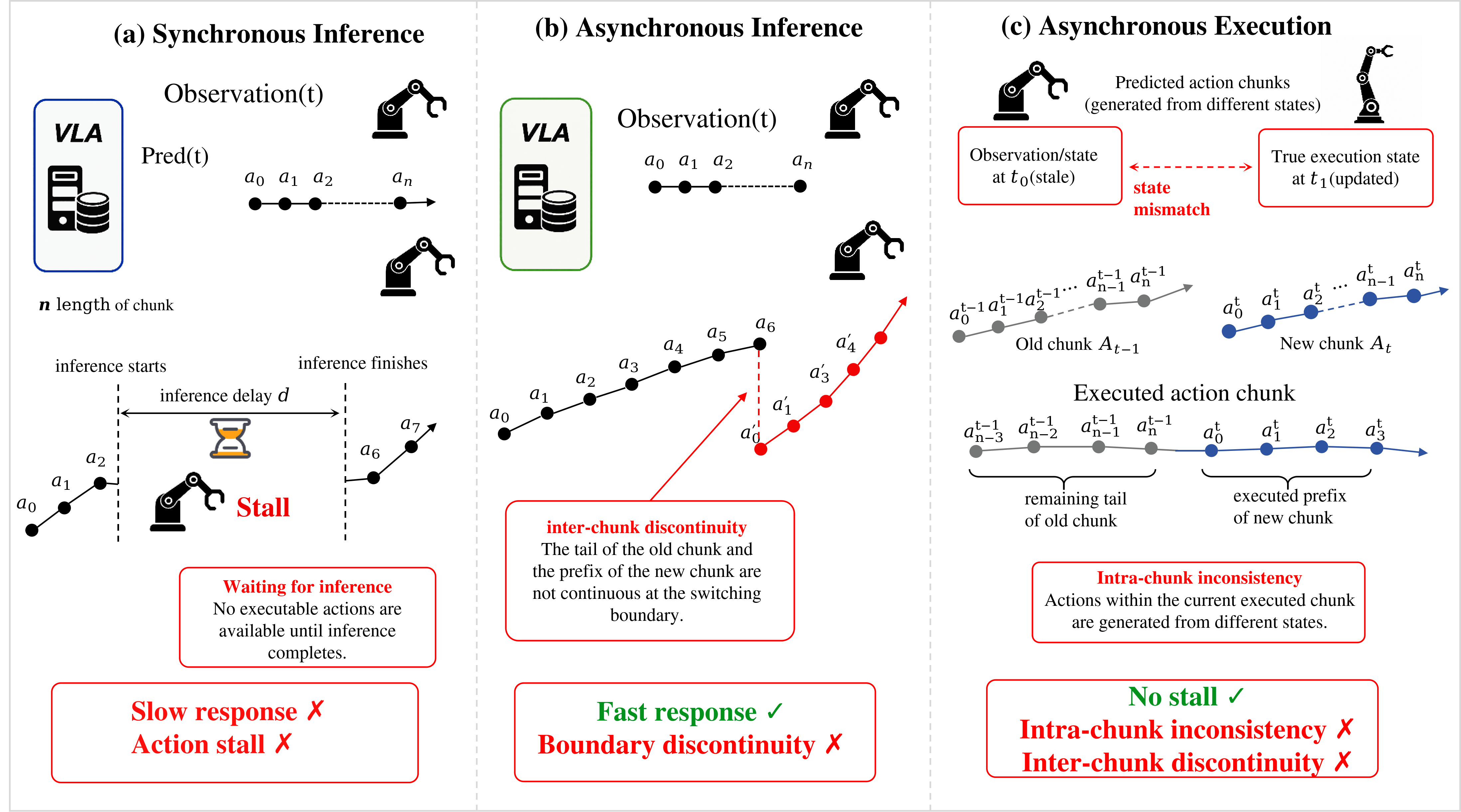}
    \caption{\textbf{Time-scale mismatch between VLA inference and robot control.}}
    \label{fig:motivation}
\end{figure}
\clearpage
\begingroup
\setstretch{1.0}
\hypersetup{linktoc=all}
\pdfbookmark[1]{Contents}{contents}
\tableofcontents
\endgroup
\clearpage
\input{sections/01_introduction_en}
\input{sections/03_latency_en}
\input{sections/04_two_stage_en}

\input{sections/05_framework_en}
\input{sections/06_experiments_en}

\FloatBarrier
\input{sections/02_related_work_en}
\input{sections/07_conclusion_en}

\clearpage
\bibliographystyle{magicw0ref}
\bibliography{references,references_root}

\clearpage
\appendix
\phantomsection
\addcontentsline{toc}{section}{Appendix}
\centerline{\Large\sffamily\bfseries Appendix}
\vspace{-9pt}
\begingroup
\let\magicmainaddcontentsline\addcontentsline
\renewcommand{\addcontentsline}[3]{%
    \ifstrequal{#1}{toc}{%
        \ifstrequal{#2}{section}{%
            \magicmainaddcontentsline{#1}{subsection}{#3}%
        }{%
            \ifstrequal{#2}{subsection}{%
                \magicmainaddcontentsline{#1}{subsubsection}{#3}%
            }{\magicmainaddcontentsline{#1}{#2}{#3}}%
        }%
    }{\magicmainaddcontentsline{#1}{#2}{#3}}%
}
\small
\setstretch{1.0}
\setlength{\abovedisplayskip}{6pt}
\setlength{\belowdisplayskip}{6pt}
\setlength{\abovedisplayshortskip}{4pt}
\setlength{\belowdisplayshortskip}{4pt}
\setcounter{figure}{0}
\renewcommand{\thefigure}{S\arabic{figure}}
\setcounter{table}{0}
\renewcommand{\thetable}{S\arabic{table}}
\input{sections/appendices_en}

\input{sections/appendix_e_execution_en}
\endgroup
\end{document}

%% file: preamble.tex
\usepackage{fix-cm}
\usepackage{fontspec}
\usepackage[english]{babel}
\newfontfamily\magictitlefont{texgyreheros-regular.otf}[
    Ligatures=TeX,BoldFont=texgyreheros-bold.otf,
    ItalicFont=texgyreheros-italic.otf,
    BoldItalicFont=texgyreheros-bolditalic.otf]
\usepackage[a4paper,margin=2.45cm]{geometry}
\usepackage{microtype}
\usepackage{setspace}
\usepackage{graphicx}
\usepackage{float}
\usepackage{subcaption}
\usepackage{booktabs}
\usepackage{array}
\usepackage{multirow}
\usepackage{tabularx}
\usepackage{makecell}
\usepackage{amsmath,amssymb,bm,mathtools}
\usepackage{algorithm}
\usepackage{algpseudocode}
\usepackage{capt-of}
\usepackage{xcolor}
\usepackage[most]{tcolorbox}
\definecolor{MagicBlue}{HTML}{2D68D8}
\definecolor{MagicAbstractBg}{HTML}{F2FAFE}
\definecolor{MagicAbstractBorder}{HTML}{C3E2F2}
\usepackage{enumitem}
\setlist{nosep,leftmargin=1.6em}
\usepackage[numbers,sort&compress]{natbib}
\usepackage{ragged2e}
\usepackage{xurl}
\usepackage[titles]{tocloft}
\usepackage{titlesec}
\usepackage[bottom]{footmisc}
\usepackage[colorlinks=true,linkcolor=MagicBlue,citecolor=MagicBlue,urlcolor=MagicBlue]{hyperref}
\usepackage{placeins}
\usepackage{wrapfig}
\usepackage{flafter}
\usepackage{needspace}
\usepackage{etoolbox}

\newcommand{\tablefont}{\fontsize{8.5}{11}\selectfont}
\renewcommand{\arraystretch}{1.12}
\AtBeginDocument{%
    \setlength{\abovedisplayskip}{6pt plus 2pt minus 1pt}%
    \setlength{\belowdisplayskip}{6pt plus 2pt minus 1pt}%
    \setlength{\abovedisplayshortskip}{4pt plus 1pt minus 1pt}%
    \setlength{\belowdisplayshortskip}{4pt plus 1pt minus 1pt}%
}
\titleformat{\section}{\normalfont\color{black}\fontsize{12}{14.4}\selectfont\bfseries}{\thesection}{0.75em}{}
\titlespacing*{\section}{0pt}{18pt plus 2pt minus 2pt}{8pt}
\titleformat{\subsection}{\normalfont\fontsize{11}{13.2}\selectfont\bfseries}{\thesubsection}{0.75em}{}
\titlespacing*{\subsection}{0pt}{12pt plus 2pt minus 1pt}{6pt}
\titleformat{\subsubsection}{\normalfont\fontsize{10}{12}\selectfont\bfseries}{\thesubsubsection}{0.75em}{}
\titlespacing*{\subsubsection}{0pt}{10pt plus 1pt minus 1pt}{4pt}
\titleformat{\paragraph}[runin]{\normalfont\bfseries}{}{0pt}{}
\titlespacing*{\paragraph}{0pt}{4pt}{0.5em}
\newtcolorbox{magicabstractbox}{
    enhanced,
    colback=MagicAbstractBg,
    colframe=MagicAbstractBorder,
    boxrule=0.45pt,
    arc=4mm,
    left=15pt,right=15pt,top=12pt,bottom=13pt,
    before skip=12pt,after skip=0pt
}
\renewenvironment{abstract}{%
    \begin{magicabstractbox}%
    \rmfamily\fontsize{10}{12}\selectfont\setstretch{1.1}%
    \setlength{\parskip}{0.18em}%
    {\centering\rmfamily\bfseries\fontsize{12}{14.4}\selectfont\color{black} Abstract\par}\vspace{6pt}\noindent\ignorespaces
}{%
    \end{magicabstractbox}%
}

\newcolumntype{Y}{>{\centering\arraybackslash}X}
\newcolumntype{L}{>{\raggedright\arraybackslash}X}
\newcommand{\figref}[1]{Fig.~\ref{#1}}
\newcommand{\tabref}[1]{Table~\ref{#1}}
\AtBeginEnvironment{thebibliography}{\RaggedRight}
\apptocmd{\bibsection}{\phantomsection\addcontentsline{toc}{section}{\refname}}{}{}
\pretocmd{\section}{\FloatBarrier\Needspace{6\baselineskip}}{}{}
\pretocmd{\subsection}{\Needspace{4\baselineskip}}{}{}
\pretocmd{\subsubsection}{\Needspace{3\baselineskip}}{}{}
\hypersetup{pdftitle={Toward Real-Time VLAs: Stage-Aware Two-Step Flow Denoising and System-Level Evaluation},pdfauthor={Di Wu, Rongtian Shen, Ping Liu, Yan Shen, Zhenhan Yin, Shun Zuo, Xuhua Chen, He Zheng, Lingfeng Zhang, Jiangning Zhang, Tao Zhang},pdflang={en-US}}

%% file: sections/00_abstract_en.tex
\begin{abstract}
Vision-language-action (VLA) models face a timing gap between low-rate inference and high-rate robot execution.
We characterize this gap through end-to-end latency measurements of model inference and the robot execution chain.
Repeated Flow Matching denoising contributes substantially to inference cost, while robot-side delays mainly arise from perception acquisition, communication scheduling, and physical response.
Analysis of the velocity field shows relatively stable magnitude and direction in early integration, followed by stronger directional correction near the terminal steps.
Based on this stage heterogeneity, we propose \textbf{two-stage non-uniform denoising}, reducing the number of steps from 10 to 2 and model-inference time from 61.557~ms to 21.956~ms.
We also develop a \textbf{distributed real-time VLA framework} with independent inference, action-publication, and robot-control rates, modular observation acquisition, and action-provenance logging.
Using $\pi_{0.5}$ as the baseline, we evaluate six real-time execution methods on a long-horizon physical garment-folding task.
Legato performs best overall among training-based methods, while Temporal Smoothing leads among training-free methods; both perform strongly in task success, completion time, action continuity, and acceleration smoothness.
Combining two-step denoising with representative execution methods substantially reduces inference cost with a small reduction in task performance.
These results motivate joint optimization of model-inference efficiency and robot-system timing.
\par\smallskip
\noindent\textbf{Project page:} \url{https://embodied.magiclab.top/works/inference/index.html}
\par\noindent\textbf{GitHub:} \url{https://github.com/MagiclabRobotics/Inference}
\end{abstract}

%% file: sections/01_introduction_en.tex
\section{Introduction}
\label{sec:introduction}

Vision--language--action (VLA) models jointly model visual observations, language instructions, and robot actions, providing a path toward general-purpose robot policies with cross-task generalization~\cite{rt2,openvla,pi0,smolvla}.
However, their computational complexity creates a substantial time-scale mismatch with the high-rate continuous control required by physical robots.
Under the action-chunk execution paradigm~\cite{act}, the policy periodically predicts a future action sequence from the current observation, while model inference, perception, communication, scheduling, and robot response all introduce delays.
When policy updates cannot reflect the current robot and environment state in time, actions generated from historical observations may continue to affect an evolving system, misaligning perception, inference, and execution.

VLA systems commonly use asynchronous inference and execution to reduce the control pauses introduced by synchronous inference: the robot executes the current action chunk while the next is generated in parallel.
As shown in \figref{fig:motivation}, this overlaps computation and execution but does not eliminate temporal misalignment caused by low-rate model updates.
A new chunk is conditioned on the observation available when inference starts, yet the robot state has changed by the time it takes control, potentially causing inter-chunk discontinuity.
Meanwhile, the robot continues executing actions predicted from historical observations during generation, causing intra-chunk inconsistency between current perception and ongoing execution~\cite{remac}.
Camera exposure and readout, proprioceptive feedback, communication and task scheduling, and physical response introduce additional system-level timing offsets~\cite{realtimev2}.
Real-time VLA is therefore an end-to-end timing problem jointly determined by model computation and the robot execution chain.

Existing real-time VLA research addresses these problems at three levels.
The first improves model-side action-generation efficiency through distillation, fewer sampling steps, or modified sampling procedures for diffusion and Flow Matching policies~\cite{snapflow,faster}.
Low-dimensional action parameterizations, such as spline control points, further reduce generation complexity while introducing smoothness and motion-feasibility priors~\cite{beast,bspline,spline,abpolicy}.
The second addresses action continuity and state alignment during asynchronous execution.
Executed or committed action prefixes condition generation during inference or training to reduce mode changes and boundary discontinuities~\cite{rtc,ttrtc,legato}.
Other methods revise actions using recent observations or predict the robot state at its future execution time, reducing observation--action and prediction--execution mismatch~\cite{vlash,remac,a2c2}.
The third jointly optimizes perception, computation, and execution through sensor timing calibration, multimodal synchronization, trajectory retiming, and local tracking control~\cite{realtimev2}.

Despite these advances, three issues remain.
First, studies use different base models, robot platforms, tasks, inference hardware, and runtime configurations, making results difficult to compare in the absence of a unified deployment framework and evaluation standard.
Second, the roles of different integration stages in the widely used Flow Matching action-generation process remain insufficiently characterized.
Third, studies commonly optimize individual components without unified end-to-end latency measurement covering model inference, perception, and robot response, making it difficult to distinguish the effects of different latency sources.

Using $\pi_{0.5}$~\cite{pi05} as the baseline, we first measure latency at both the model and robot levels to identify the main end-to-end sources.
We then analyze visual--language conditioning and Flow Matching generation costs, together with velocity-field magnitude and direction across integration steps.
These observations motivate a two-stage non-uniform denoising strategy that reduces the number of function evaluations (NFE) from $10$ to $2$ through non-uniform integration times, substantially improving model-inference efficiency.
For system-side latency, we develop a distributed, thread-decoupled VLA inference and execution framework with modular inference, action publication, temporal alignment, and local control.
Using common training data, a shared robot platform, and standardized runtime configurations, we evaluate Naive Asynchronous Execution, Temporal Smoothing, Inference-time RTC, Training-time RTC, Legato, and VLASH on a long-horizon bimanual garment-folding task.
We also evaluate the proposed two-stage strategy in combination with representative real-time execution methods.

Our main contributions are:
\begin{enumerate}
    \setlength{\itemsep}{0.25em}
    \setlength{\parsep}{0pt}
    \setlength{\parskip}{0pt}
    \item \textbf{A systematic procedure for measuring model-side and robot-side latency.}
    Using $\pi_{0.5}$, we analyze visual--language conditioning and Flow Matching generation, and measure camera timestamp offset, image readout, proprioceptive feedback, and control response to quantify end-to-end VLA timing.
    \item \textbf{A two-stage non-uniform Flow Matching denoising strategy.}
    Stepwise analysis reveals stage heterogeneity in Flow Matching denoising.
    Non-uniform integration combines standard Flow long-interval generation with SnapFlow short-interval terminal refinement, reducing NFE from $10$ to $2$ and mean model-inference time from $61.557$ ms to $21.956$ ms, a $2.804\times$ speedup.
    \item \textbf{An open-source distributed real-time VLA inference and execution framework.}
    Modular, thread-decoupled components jointly manage observation acquisition, policy inference, action buffering, temporal alignment, local control, and runtime logging.
    Independent inference, action-publication, and robot-control rates provide a unified, traceable basis for deploying and evaluating real-time VLA methods.
\end{enumerate}

%% file: sections/03_latency_en.tex
\section{Sources of End-to-End Latency and Model-Inference Bottlenecks}
\label{sec:latency_bottlenecks}

Real-time VLA methods improve action generation, asynchronous scheduling, action-chunk handover, and execution-time correction.
However, model inference rate alone cannot explain the temporal offset accumulated before an action physically takes effect~\cite{realtimev2}.
We distinguish two classes of latency: delays in perception, communication, scheduling, and physical execution; and visual--language computation and Flow Matching action generation within the model.

\begin{table}[!htbp]
    \centering
    \caption{\textbf{Measured latency components.}}
    \label{tab:latency_bottleneck_summary}
    \tablefont
    \setlength{\tabcolsep}{5pt}
    \renewcommand{\arraystretch}{1.12}
    \begin{tabularx}{\textwidth}{@{}l@{\hspace{8pt}}>{\raggedright\arraybackslash}p{96pt}@{\hspace{8pt}}c@{\hspace{8pt}}L@{}}
        \toprule
        \textbf{Layer} & \textbf{Latency component} & \textbf{Measured value} & \textbf{Interpretation} \\
        \midrule
        Perception & Primary-camera\newline timestamp offset & $17.7\pm1.8$ ms & Offset of the RealSense internal timestamp relative to actual image~acquisition \\
        Perception & Primary-camera\newline image readout & $32.8\pm3.3$ ms & Input/output (I/O) delay from the RealSense timestamp to image receipt by the host process \\
        State & Proprioceptive feedback & $26.4\pm6.2$ ms & Delay from the physical robot state to host observation of the corresponding software development kit (SDK) feedback \\
        Execution & Command-to-motion\newline response & $15.3\pm5.4$ ms & Delay from command publication until physical motion follows the target phase \\
        Model & Standard Flow, 10 NFE & $61.6\pm6.2$ ms & Latency statistics from $620$ inference calls \\
        \bottomrule
    \end{tabularx}
\end{table}

\subsection{End-to-End Timeline and Non-Model Latency}

A closed-loop update begins with image formation and a change in robot state, followed by image readout, state feedback, input preprocessing, request transmission, policy inference, result transmission, action buffering and publication, and finally physical response.
The interval from host receipt of an observation to availability of a new action chunk is
\begin{equation}
    T_{\mathrm{avail}}
    =T_{\mathrm{pre}}+T_{\mathrm{up}}+T_{\mathrm{model}}
    +T_{\mathrm{down}}+T_{\mathrm{queue}},
    \label{eq:action_available_latency}
\end{equation}
where $T_{\mathrm{pre}}$ denotes input preprocessing, $T_{\mathrm{up}}$ and $T_{\mathrm{down}}$ denote request upload and action-chunk download, and $T_{\mathrm{queue}}$ includes service queueing, action buffering, and control-period quantization.
Equation~\eqref{eq:action_available_latency} excludes perception latency accumulated before host receipt and physical response after command publication.
The visual information used when an action takes effect is therefore also subject to the camera timestamp offset $t_{\mathrm{camera}}$ and image-readout latency $t_{\mathrm{readout}}$.
Proprioceptive state carries feedback latency $t_{\mathrm{proprio}}$, and the output path includes motion-response latency $t_{\mathrm{motion}}$.

\begin{figure}[H]
    \centering
    \includegraphics[trim=0bp 60bp 0bp 20bp,clip,width=0.88\textwidth]{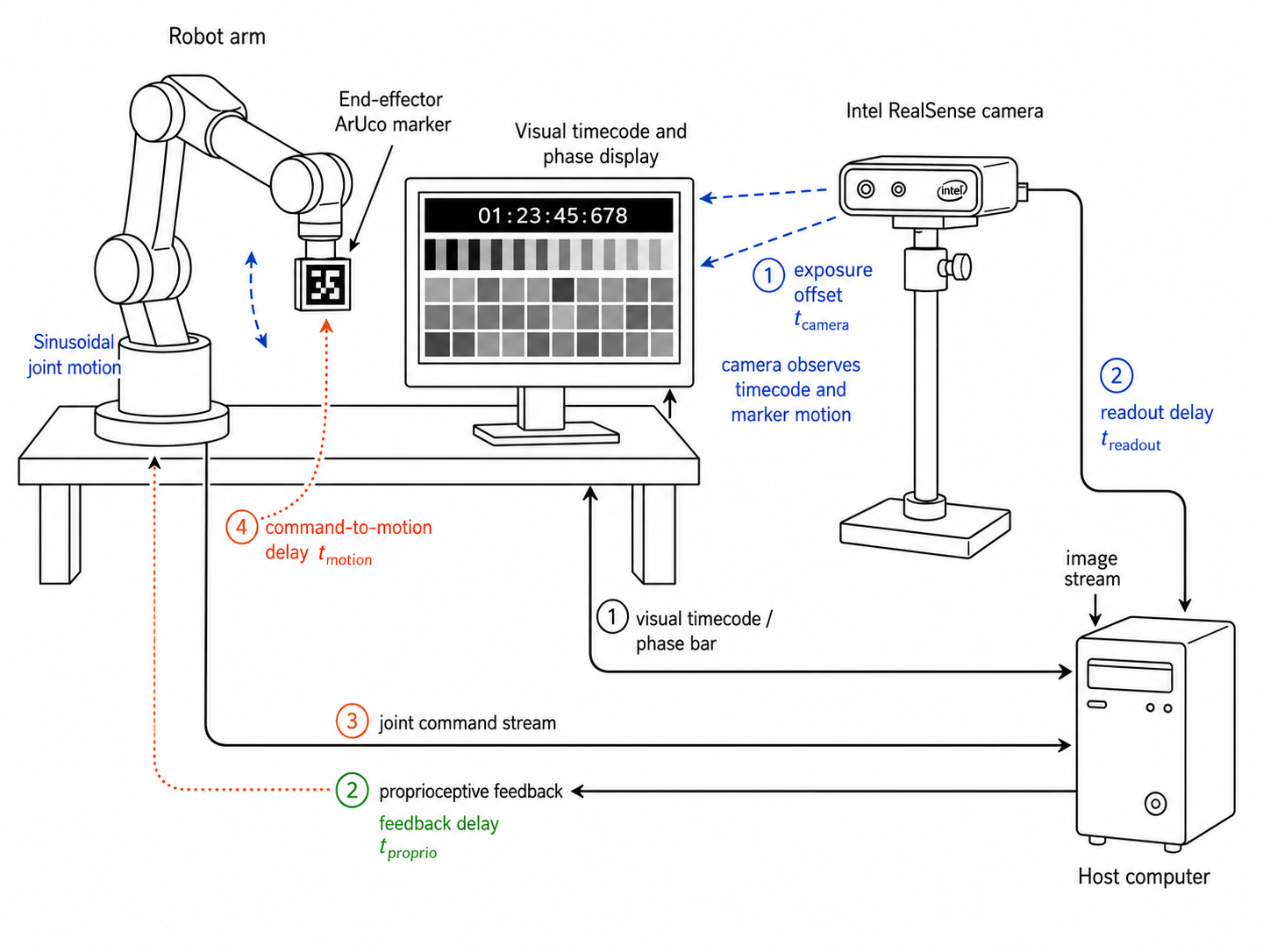}
    \caption{\textbf{Joint system-latency calibration apparatus and signal-acquisition workflow.} The RealSense camera records a visual time code on the display and an end-effector ArUco marker, while the host logs image timestamps, receipt times, joint commands, and proprioceptive feedback. Timestamp differences and phase relationships among periodic signals estimate the camera timestamp offset, image-readout latency, motion-response latency, and proprioceptive-feedback latency.}
    \label{fig:latency_calibration}
\end{figure}

We measure these components using the joint calibration apparatus in \figref{fig:latency_calibration}.
The robot executes small-amplitude periodic sinusoidal joint motions while a display presents a host time code and phase bar.
A RealSense camera records both the visual time code and an end-effector ArUco marker; the host logs camera timestamps, image receipt times, joint commands, and SDK feedback.
The difference between host receipt time and the RealSense frame timestamp estimates image-readout latency, while the display time code recovers the actual image-acquisition time.
Phase or correlation relationships among periodic commands, visual marker trajectories, and state feedback separate motion-response latency from proprioceptive-feedback latency.
Calibration uses visual acquisition at 30 frames/s and state recording at 100 Hz, with repeated measurements across joints and motion frequencies.

\tabref{tab:latency_bottleneck_summary} summarizes the principal latency components measured on the physical robot system.
Beyond model inference, perception, state feedback, and physical execution each incur non-negligible delays.
The primary-camera timestamp offset and image-readout latency are $17.690$ ms and $32.817$ ms, respectively; proprioceptive-feedback latency is $26.4\pm6.2$ ms, and command-to-motion response latency is $15.3\pm5.4$ ms.
Thus, the observations available to the controller are inherently misaligned in time with the physical robot state.
Faster model inference alone cannot eliminate this mismatch.

\FloatBarrier
\subsection{Model-Inference Time Decomposition}

For a VLA with a Flow Matching action head, one sampling call can be expressed as
\begin{equation}
    T_{\mathrm{model}}(N)
    =T_{\mathrm{cond}}
    +\sum_{k=1}^{N}T_{\mathrm{step}}^{(k)}
    +T_{\mathrm{post}},
    \label{eq:model_latency_decomposition}
\end{equation}
where $T_{\mathrm{cond}}$ includes image preprocessing, visual--language prefix encoding, and condition caching, whose cost does not scale proportionally with the number of denoising steps.
$T_{\mathrm{step}}^{(k)}$ is the cost of the $k$th Flow Matching sampling step, including action-head evaluation and its numerical integration update; $T_{\mathrm{post}}$ covers action denormalization and output transformation.
Model-inference time therefore comprises fixed costs independent of NFE and repeated sampling costs that increase with NFE.

On an NVIDIA GeForce RTX 4090 D (24 GB) in JAX, standard 10-NFE Flow takes $61.557$ ms on average across $620$ inference calls after excluding the first just-in-time (JIT) compilation.
This result indicates substantial model-side computation in iterative Flow sampling.
The next section analyzes differences between sampling stages and introduces two-stage non-uniform time integration to reduce the NFE required for action generation.

%% file: sections/04_two_stage_en.tex
\section{Two-Stage Non-Uniform Denoising}
\label{sec:two_stage_denoising}

The measurements in Section~\ref{sec:latency_bottlenecks} show that, in addition to non-model delays in perception, state feedback, and physical execution, standard Flow action generation requires repeated velocity-field evaluations.
Reducing the number of function evaluations during action generation is therefore a direct route to lower model-inference cost.
Simply reducing the number of integration steps, however, generally increases discretization error.
The key question is how to allocate a limited computational budget along Flow time.

\subsection{Stage Heterogeneity of the Denoising Velocity Field}

To characterize changes between consecutive Flow Matching iterations, we fix model parameters, language condition, and initial noise, and record complete velocity-field trajectories from standard 10-NFE inference for different observations.
For the velocity field $v_k\in\mathbb{R}^{50\times14}$ at evaluation $k$, root mean square (RMS) measures its overall magnitude in normalized action space:
\begin{equation}
    r_k=\frac{\lVert v_k\rVert_{\mathrm{F}}}{\sqrt{50\times14}},
    \label{eq:flow_velocity_rms}
\end{equation}
where $\lVert\cdot\rVert_{\mathrm{F}}$ is the Frobenius norm.
The angle between adjacent velocity fields measures the directional correction at integration step $k$:
\begin{equation}
    \theta_k=
    \arccos\!\left(
        \frac{\langle v_k,v_{k-1}\rangle_{\mathrm{F}}}
        {\lVert v_k\rVert_{\mathrm{F}}\lVert v_{k-1}\rVert_{\mathrm{F}}}
    \right)\frac{180^\circ}{\pi},
    \quad k=2,\ldots,10.
    \label{eq:flow_velocity_angle}
\end{equation}

\begin{figure}[!htbp]
    \centering
    \includegraphics[width=0.92\textwidth]{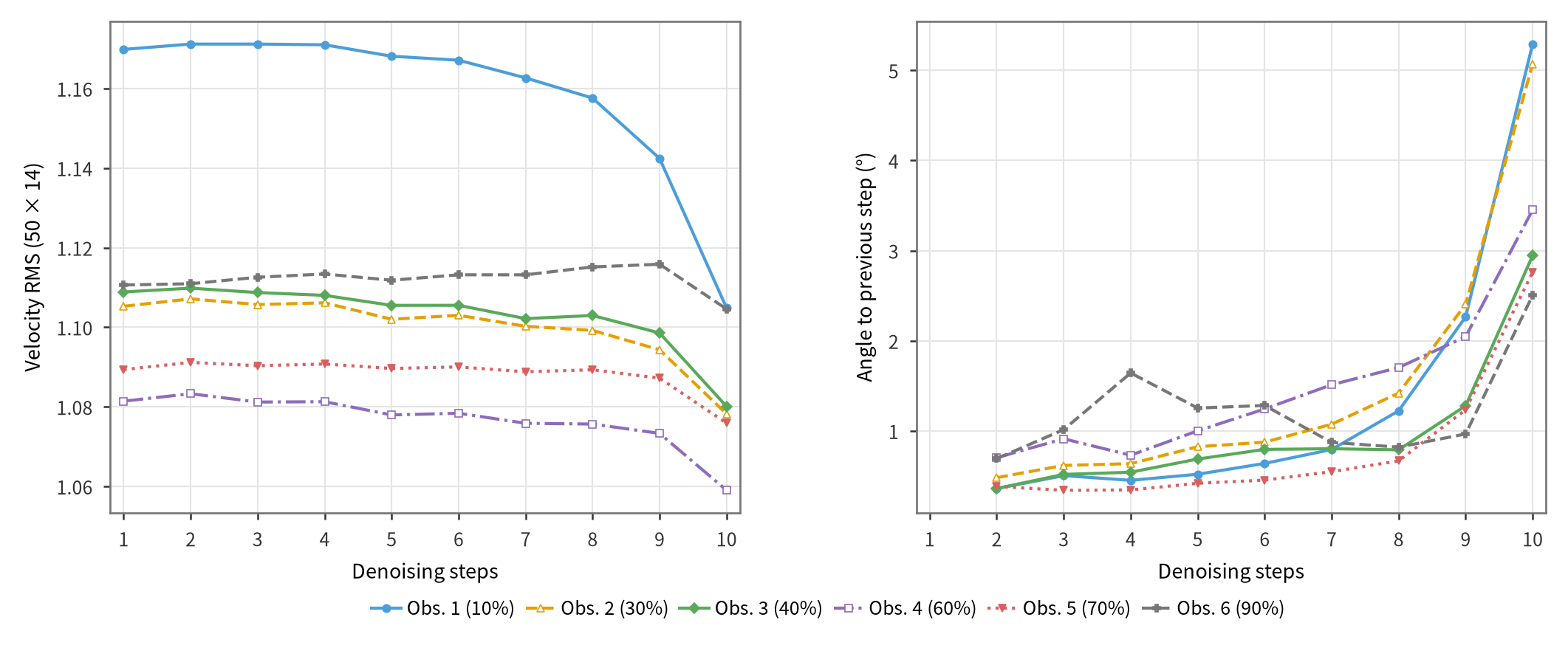}
    \caption{\textbf{Velocity-field trajectories under standard 10-NFE Flow sampling.} Changes in velocity RMS $r_k$ (left) and the angle $\theta_k$ between adjacent integration steps (right) are concentrated near the endpoint.}
    \label{fig:flow_stage_motivation}
\end{figure}

As shown in \figref{fig:flow_stage_motivation}, all six velocity RMS trajectories remain broadly stable during the early and intermediate stages, with small angles between adjacent velocity fields.
Near the denoising endpoint, most trajectories show more pronounced magnitude changes, and all six angle trajectories rise at the final step.
These observations indicate that network evaluations contribute unevenly to action generation: early and intermediate updates progress along similar velocity fields, whereas terminal updates concentrate magnitude adjustments and directional corrections.
This stage difference motivates long-interval coarse generation followed by short-interval terminal refinement.

Accordingly, we allocate the two network evaluations non-uniformly along Flow time.
The first spans a longer interval to generate the main action chunk, and the second refines the prediction over a short interval near the endpoint.

\FloatBarrier
\subsection{Flow Matching Action Generation and Standard Sampling}

Let $c$ denote the condition formed by three images, a language instruction, and robot state.
Let $x_t\in\mathbb{R}^{H\times d}$ be the normalized action chunk at time $t$, where $t=1$ denotes Gaussian noise and $t=0$ denotes the target action.
The Flow Matching~\cite{flowmatching} action expert learns the conditional velocity field
\begin{equation}
    \frac{\mathrm{d}x_t}{\mathrm{d}t}=v_\theta(x_t,t;c),
    \qquad x_1\sim\mathcal{N}(0,I).
    \label{eq:flow_ode}
\end{equation}
For a decreasing time grid $1=t_0>t_1>\cdots>t_N=0$, standard Euler sampling applies
\begin{equation}
    x_{t_{k+1}}=x_{t_k}+(t_{k+1}-t_k)
    v_\theta(x_{t_k},t_k;c).
    \label{eq:flow_euler}
\end{equation}
The action expert therefore performs $N$ function evaluations.
Section~\ref{sec:latency_bottlenecks} shows that this repeated path grows with NFE and is a direct target for reducing model-side cost.

\subsection{Two-Stage Non-Uniform Partition}

We partition the integration interval into $[1,\tau]$ and $[\tau,0]$, fixing $\tau=0.3$ in the current implementation.
The first stage spans an interval of length $0.7$ for coarse action generation.
The second spans the final $0.3$ interval to restore action details and correct terminal residuals.

\begin{figure}[!htbp]
    \centering
    \includegraphics[trim=0bp 12bp 0bp 5bp,clip,width=0.90\textwidth]{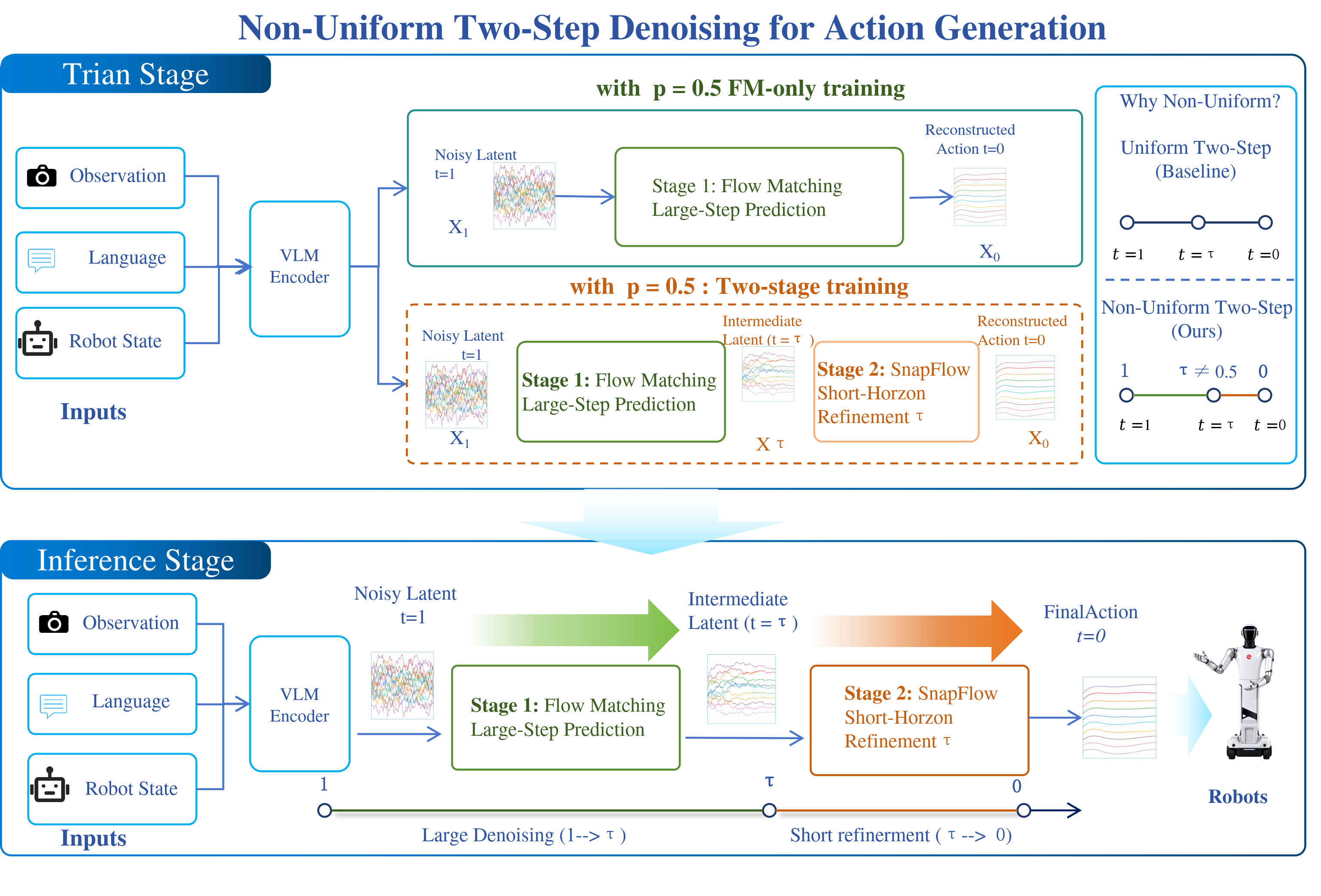}
    \caption{\textbf{Training and inference pipeline for two-stage non-uniform action denoising.} Training selects the standard Flow Matching branch or the two-stage branch with equal probability. The latter applies a large Flow update followed by short-interval mean-velocity prediction for terminal refinement. Inference reuses the same visual--language--proprioceptive condition and evaluates the action expert only twice along the non-uniform schedule $1\rightarrow\tau\rightarrow0$. The right side compares uniform two-step sampling with the proposed asymmetric schedule.}
    \label{fig:two_stage_denoising_overview}
\end{figure}

As shown in \figref{fig:two_stage_denoising_overview}, the first stage evaluates a local Flow velocity at $t=1$ and applies a large Euler step.
The second takes the predicted state $x_\tau$, current time $\tau$, and target time $0$ as inputs to predict the mean velocity over the terminal interval.
Both velocities are predicted by the same action expert and distinguished by target time $s$.
Define $\mathcal{V}_\theta(x,t\!\rightarrow\!s;c)=\operatorname{clip}(v_\theta(x,t,s;c),-C,C)$, where $C=20$.
Here, $s=t$ denotes local velocity prediction, and $s=0$ denotes cross-interval velocity prediction directly to the denoising endpoint.
The updates are
\begin{align}
    v^{(1)} &= \mathcal{V}_\theta(x_1,1\!\rightarrow\!1;c),
    &x_\tau &= x_1-(1-\tau)v^{(1)}, \nonumber\\
    \bar v^{(2)} &= \mathcal{V}_\theta(x_\tau,\tau\!\rightarrow\!0;c),
    &x_0 &= x_\tau-\tau\bar v^{(2)},
    \label{eq:two_stage_inference}
\end{align}
with no teacher branch during deployment and only two action-expert evaluations.
The stages share condition $c$ and action-expert parameters, but each still executes a complete stage computation.
A fivefold reduction in NFE therefore does not imply a fivefold reduction in execution time.

\FloatBarrier
\subsection{SnapFlow Terminal Refinement and Joint Training}

In the proposed two-stage framework, the first stage uses standard Flow Matching for long-interval action generation.
The second directly adopts the shortcut refinement mechanism of SnapFlow~\cite{snapflow}, restricted to the short terminal interval $[\tau,0]$.
Unlike one-step or few-step compression over the full Flow interval, our method retains standard Flow generation in the first stage and confines SnapFlow to the terminal interval.

\begin{algorithm}[!htbp]
    \caption{\textbf{Two-stage non-uniform denoising: training and deployment inference.}}
    \label{alg:two_stage_denoising}
    \small
    \begin{algorithmic}[1]
        \Require Condition $c$, training action chunk $x_0$,
        $\tau=0.3$, $\lambda_{\mathrm{sc}}=0.1$
        \Ensure Updated parameters $\theta$; generated action chunk $\hat{x}_0$ at deployment
        \Statex \textbf{Training:}
        \State Sample $x_1\sim\mathcal{N}(0,I)$; set $v^*\gets x_1-x_0$
        \State Sample $b\in\{0,1\}$ with equal probability
        \If{$b=1$} \Comment{Standard Flow Matching}
            \State Sample $t\sim\mathcal{U}(0,1)$ for each example
            \State $x_t\gets(1-t)x_0+t x_1$
            \State $\hat v\gets\mathcal{V}_\theta(x_t,t\!\rightarrow\!t;c)$
            \State $\mathcal{L}\gets\operatorname{MSE}(\hat v,v^*)$
        \Else \Comment{Non-uniform two-stage branch}
            \Statex \hspace{\algorithmicindent}\textit{Stage 1: Standard Flow large-step prediction}
            \State $v^{(1)}\gets\mathcal{V}_\theta(x_1,1\!\rightarrow\!1;c)$
            \State $x_\tau\gets x_1-(1-\tau)v^{(1)}$
            \Statex \hspace{\algorithmicindent}\textit{Stage 2: SnapFlow short-horizon refinement}
            \State $\tilde{x}_\tau\gets\operatorname{sg}(x_\tau)$
            \State $v_a\gets\operatorname{sg}\!\left(\mathcal{V}_\theta(
            \tilde{x}_\tau,\tau\!\rightarrow\!\tau;c)\right)$
            \State $x_{\tau/2}\gets\tilde{x}_\tau-\frac{\tau}{2}v_a$
            \State $v_b\gets\operatorname{sg}\!\left(\mathcal{V}_\theta(
            x_{\tau/2},\tau/2\!\rightarrow\!\tau/2;c)\right)$
            \State $\bar v^{*}_{\tau\rightarrow0}\gets(v_a+v_b)/2$
            \State $\bar v^{(2)}\gets\mathcal{V}_\theta(\tilde{x}_\tau,\tau\!\rightarrow\!0;c)$
            \State $\mathcal{L}\gets\operatorname{MSE}(v^{(1)},v^*)$
            \Statex \hspace{\algorithmicindent}$\phantom{\mathcal{L}\gets{}}+
            \lambda_{\mathrm{sc}}\operatorname{MSE}(\bar v^{(2)},\bar v^{*}_{\tau\rightarrow0})$
        \EndIf
        \State Update parameters $\theta$ using $\nabla_\theta\mathcal{L}$
        \Statex \textbf{Inference:}
        \State Sample $x_1\sim\mathcal{N}(0,I)$
        \State $x_\tau\gets x_1-(1-\tau)\mathcal{V}_\theta(x_1,1\!\rightarrow\!1;c)$
        \State $\hat{x}_0\gets x_\tau-\tau\mathcal{V}_\theta(x_\tau,\tau\!\rightarrow\!0;c)$
        \State \Return $\hat{x}_0$ \Comment{Two action-expert evaluations}
    \end{algorithmic}
\end{algorithm}

The second stage follows SnapFlow in constructing cross-interval mean-velocity supervision from a shortcut target, an idea originating in Shortcut Models~\cite{shortcutmodels}.
Starting from the actual first-stage prediction $x_\tau$, we apply stop-gradient to obtain $\tilde{x}_\tau=\operatorname{sg}(x_\tau)$, preventing the terminal refinement loss from backpropagating through the first-stage long-interval prediction.
Two local velocity evaluations within $[\tau,0]$ construct the supervision target:
\begin{align}
    v_a &= \operatorname{sg}\!\left(
    \mathcal{V}_\theta(\tilde{x}_\tau,\tau\!\rightarrow\!\tau;c)
    \right),
    &x_{\tau/2} &= \tilde{x}_\tau-\frac{\tau}{2}v_a, \nonumber\\
    v_b &= \operatorname{sg}\!\left(
    \mathcal{V}_\theta(x_{\tau/2},\tau/2\!\rightarrow\!\tau/2;c)
    \right),
    &\bar v^{*}_{\tau\rightarrow0} &= \frac{v_a+v_b}{2}.
    \label{eq:shortcut_teacher}
\end{align}
Here, $\operatorname{sg}(\cdot)$ denotes stop-gradient, and $\bar v^{*}_{\tau\rightarrow0}$ is the terminal-interval mean-velocity target constructed by the SnapFlow teacher path.
The second-stage student takes $\tilde{x}_\tau$ as input and directly predicts the mean velocity from time $\tau$ to endpoint $0$:
\begin{equation}
    \bar v^{(2)}
    =\mathcal{V}_\theta(\tilde{x}_\tau,\tau\!\rightarrow\!0;c).
\end{equation}

To retain both standard Flow Matching local-velocity modeling and SnapFlow cross-interval prediction, training selects the standard Flow Matching branch or the two-stage branch with equal probability at the batch level.
For the latter, the joint objective is
\begin{equation}
\mathcal{L}_{\mathrm{2stage}}=\operatorname{MSE}\!\left(v^{(1)},v^{(1)*}\right)+\lambda_{\mathrm{sc}}\operatorname{MSE}\!\left(\bar v^{(2)},\bar v^{*}_{\tau\rightarrow0}\right),
\label{eq:two_stage_training}
\end{equation}
where $\operatorname{MSE}$ averages the squared error over action-chunk elements.
The first-stage loss weight is fixed at $1.0$, and the SnapFlow terminal refinement weight is $\lambda_{\mathrm{sc}}=0.1$.

For a training action chunk $x_0$ and Gaussian noise $x_1$, standard Flow Matching uses the linear probability path
\begin{equation}
    x_t=(1-t)x_0+t x_1,
\end{equation}
with local velocity target $v^{*}=x_1-x_0$.
Thus, $v^{(1)*}=v^{*}$ in Eq.~\eqref{eq:two_stage_training}.

Our design does not redefine SnapFlow shortcut supervision.
Instead, motivated by the observed heterogeneity across Flow integration stages, it combines standard Flow long-interval generation with SnapFlow short-interval terminal refinement to form the non-uniform schedule $1\rightarrow\tau\rightarrow0$, with $\tau=0.3$ in the current implementation.
Algorithm~\ref{alg:two_stage_denoising} summarizes training and deployment inference.

%% file: sections/05_framework_en.tex
\Needspace{8\baselineskip}
\section{Distributed Real-Time Inference and Execution Framework}
\label{sec:distributed_framework}

\subsection{Overall Architecture}

\begin{figure}[!htbp]
    \centering
    \includegraphics[trim=0bp 0bp 0bp 78bp,clip,width=0.88\textwidth]{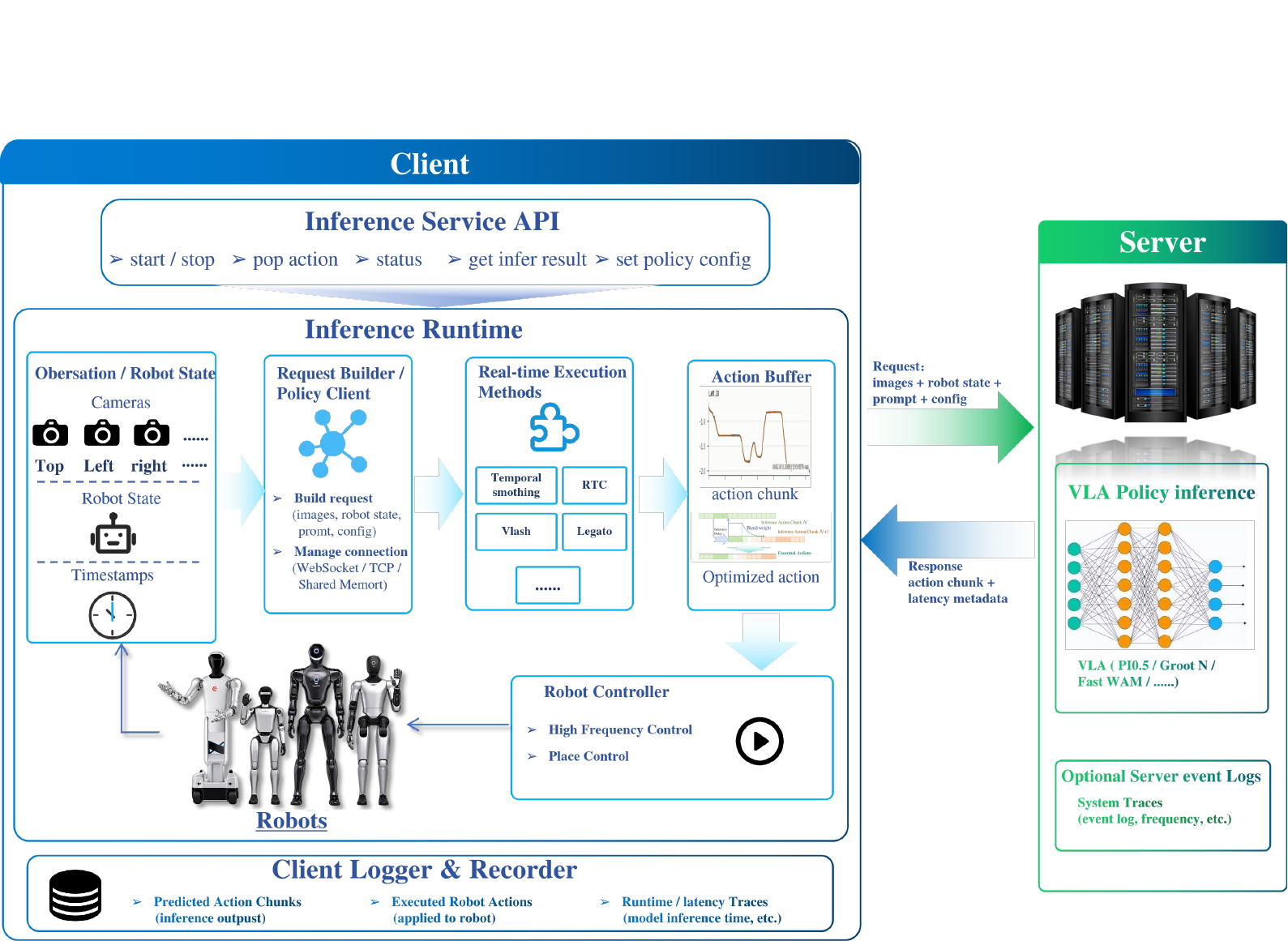}
    \caption{\textbf{Distributed real-time VLA inference and execution framework.}}
    \label{fig:realtime_vla_framework}
\end{figure}

To address the system-side latency analyzed in Section~\ref{sec:latency_bottlenecks}, we decouple robot-side observation acquisition and high-rate control from GPU-side policy computation, as shown in \figref{fig:realtime_vla_framework}.
The robot-side \texttt{Collector/RobotIO} aggregates the overhead camera, two wrist cameras, and robot state, continuously caching the latest observation.
At a configured rate, the inference thread reads a snapshot and assembles images, a language instruction, robot state, sampling steps, and execution-mode-specific latency fields into a request.
Cross-host deployment uses a Transmission Control Protocol socket or WebSocket, whereas same-host deployment can use shared memory.
The GPU policy server parses requests, transforms inputs, performs VLA inference, and returns action chunks with latency metadata.

The client uses latency estimates to determine each chunk's expected effective index, then handles handover, blending, and indexing in the action buffer.
An independent publication thread sends individual commands at the control rate, preventing low-rate inference and communication waits from blocking high-rate control.
The framework also records sensor timestamps and action provenance.
Hardware-specific clock semantics, fixed offsets, and wrist-camera timing differences require separate calibration.

\FloatBarrier
\subsection{Pluggable Strategy Integration and Runtime Tracing}

The mode manager encapsulates synchronous execution and the six real-time VLA strategies in \tabref{tab:method_taxonomy} as pluggable modules that share observation, communication, buffering, and control paths.
Logs record request submission and return, model inputs and outputs, action-chunk indices, published actions and their provenance, and request, inference, and transmission times.
These records support reconstruction of action handovers and localization of errors by system layer.
The experiments demonstrate unified integration, physical robot execution, and trajectory traceability for all six methods.

\begin{table}[!htbp]
\caption{\textbf{Layered taxonomy of six real-time VLA execution strategies.}}
\label{tab:method_taxonomy}
\centering
\tablefont
\setlength{\tabcolsep}{4pt}
\renewcommand{\arraystretch}{1.12}
\begin{tabularx}{\textwidth}{l l L L}
\toprule
\textbf{Method} & \textbf{System layer} & \textbf{Principal operation} & \textbf{Primary objective} \\
\midrule
Naive Asynchronous & Scheduling & Decouple model inference from robot execution & Remove robot waiting during inference \\
Temporal Smoothing & Action buffer & Blend the old-chunk tail with the new-chunk prefix & Suppress velocity and acceleration discontinuities \\
Inference-time RTC & Generation constraint & Fix committed actions and complete the remaining trajectory & Reduce mismatch between old and new chunks \\
Training-time RTC & Policy conditioning & Simulate latency and condition on an action prefix during training & Learn low-overhead action continuation \\
Legato & Learned generation & Learn native action-continuation dynamics & Internalize inter-chunk continuity \\
VLASH & State alignment & Propagate state to the expected execution time & Reduce prediction error caused by stale state \\
\bottomrule
\end{tabularx}
\end{table}

%% file: sections/06_experiments_en.tex
\section{Experimental Evaluation and Analysis}
\label{sec:experiments}

This section evaluates the distributed real-time execution framework and the two-stage 2-NFE denoising strategy.
Execution-layer experiments deploy six real-time VLA methods in a unified framework and use physical garment-folding results and action-continuity measurements to compare task performance and execution characteristics under practical deployment configurations.
Model-side experiments compare inference time and action error for standard 10-NFE and two-stage 2-NFE sampling under fixed offline inputs and a common timing protocol, quantifying the computational gains and error changes from fewer sampling steps.

\subsection{Distributed Inference and Execution Experiments}
\label{subsec:distributed_experiments}

To evaluate the framework's support for different real-time execution methods, we deploy six representative methods under a common software and hardware setup and compare them on a long-horizon bimanual garment-folding task.
The evaluation covers task completion, execution efficiency, and action continuity.

\subsubsection{Comparison Methods and Runtime Configurations}

We evaluate Naive Asynchronous Execution, Temporal Smoothing, Inference-time RTC, Training-time RTC, Legato, and VLASH.
\tabref{tab:method_taxonomy} summarizes their system layers and principal operations.

\figref{fig:inference_execution_modes} and \figref{fig:delay_aware_methods} use a common action-chunk timeline to illustrate inference scheduling and handover mechanisms.
Synchronous inference and Temporal Ensembling~\cite{act} serve as reference configurations for framework validation and are excluded from the unified physical comparison and its reported results.

\subsubsection{Experimental Platform and Task Setup}

The bimanual platform comprises two Agilex Piper manipulators.
Robot actions and proprioceptive state contain 12 arm-joint dimensions and two gripper dimensions.
An overhead camera and two wrist cameras each acquire $640\times480$ RGB images at 30 frames/s.

We use $\pi_{0.5}$ as the common base policy and bimanual T-shirt folding as the long-horizon physical task.
The three T-shirts differ in size and color: small purple, large pink, and medium red.
These garments introduce variation in object appearance and scale.
Each method is evaluated independently ten times on each garment, yielding 30 trials per method and 180 physical trials in total.
Inference rate, action-publication rate, and Flow Matching NFE use the best configurations identified for each method in preliminary experiments.
Final evaluation uses the same hardware platform and task conditions.

\begin{figure}[!htbp]
    \centering
    \begin{minipage}[t]{0.48\linewidth}
        \centering
        \includegraphics[trim=110bp 4bp 95bp 4bp,clip,width=\linewidth,height=4.3cm,keepaspectratio]{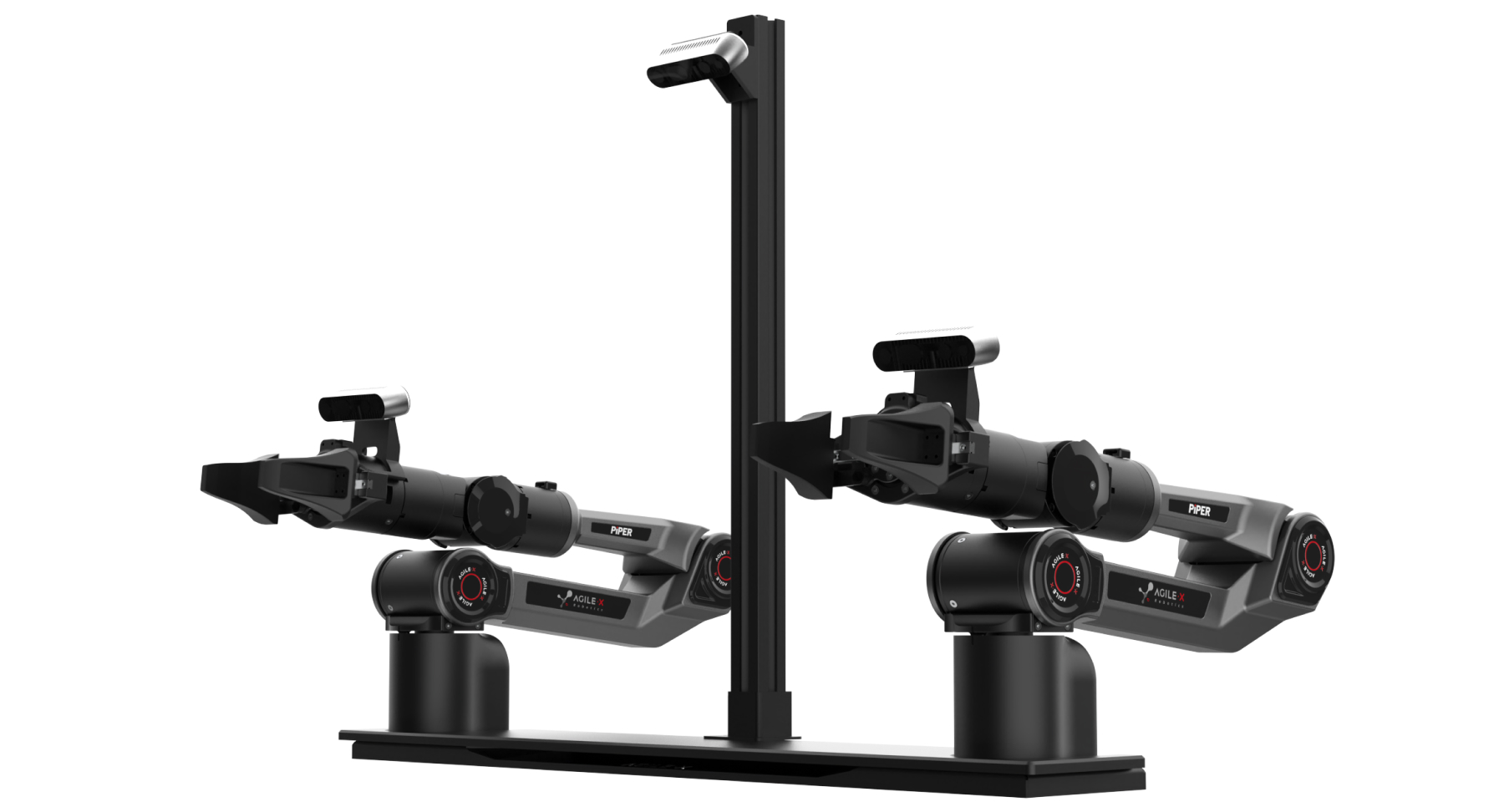}
        \caption{\textbf{Bimanual robot platform used in the experiments.}}
        \label{fig:robot_platform}
    \end{minipage}\hfill
    \begin{minipage}[t]{0.48\linewidth}
        \centering
        \includegraphics[angle=90,width=\linewidth,height=4.3cm,keepaspectratio]{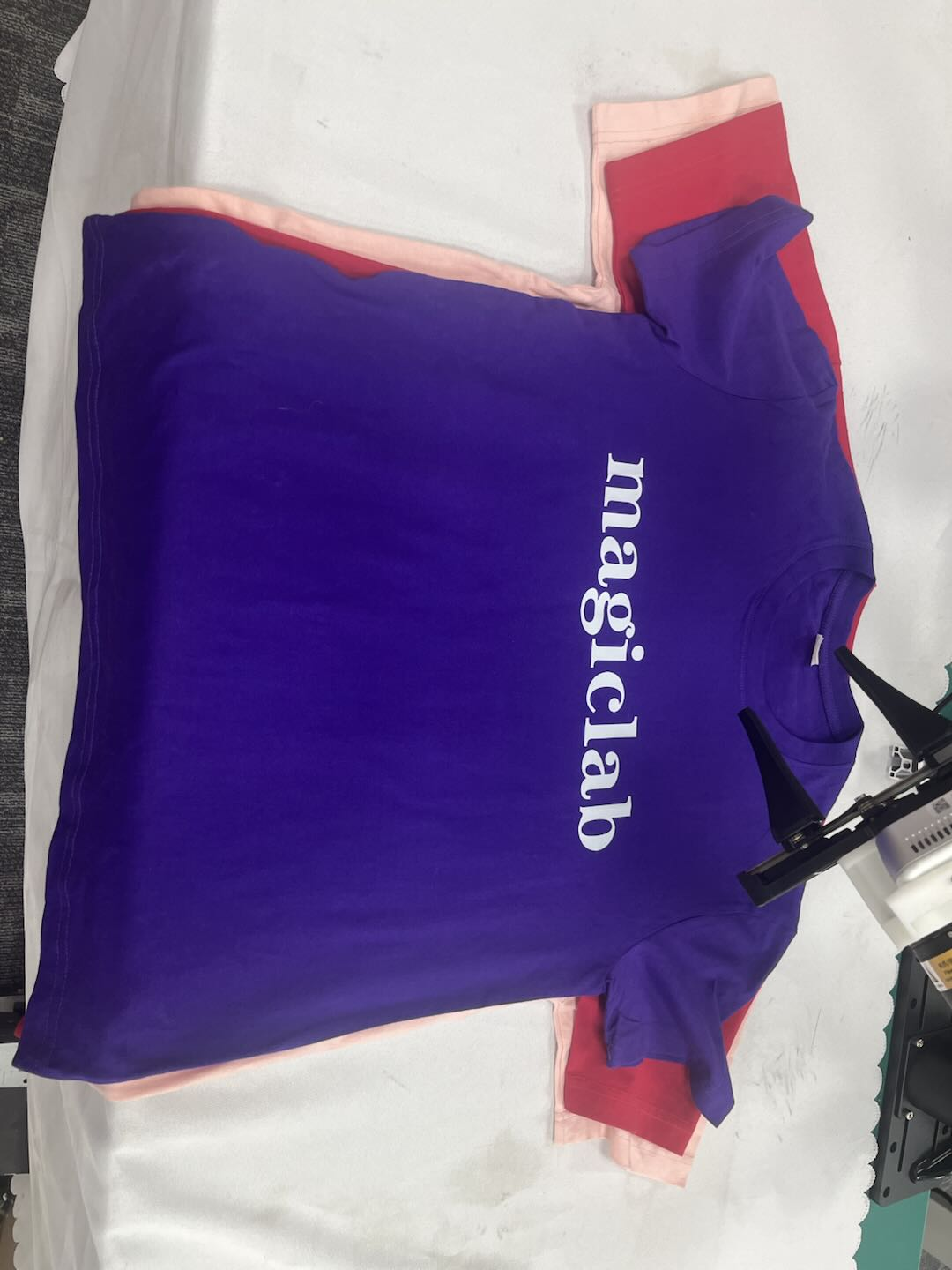}
        \caption{\textbf{Garments with different sizes and colors used in the experiments.}}
        \label{fig:garment_conditions}
    \end{minipage}
\end{figure}

\begin{figure}[!htbp]
    \centering
    \small
    \setlength{\tabcolsep}{4pt}
    \begin{tabular}{cccc}
    \includegraphics[width=0.220\textwidth]{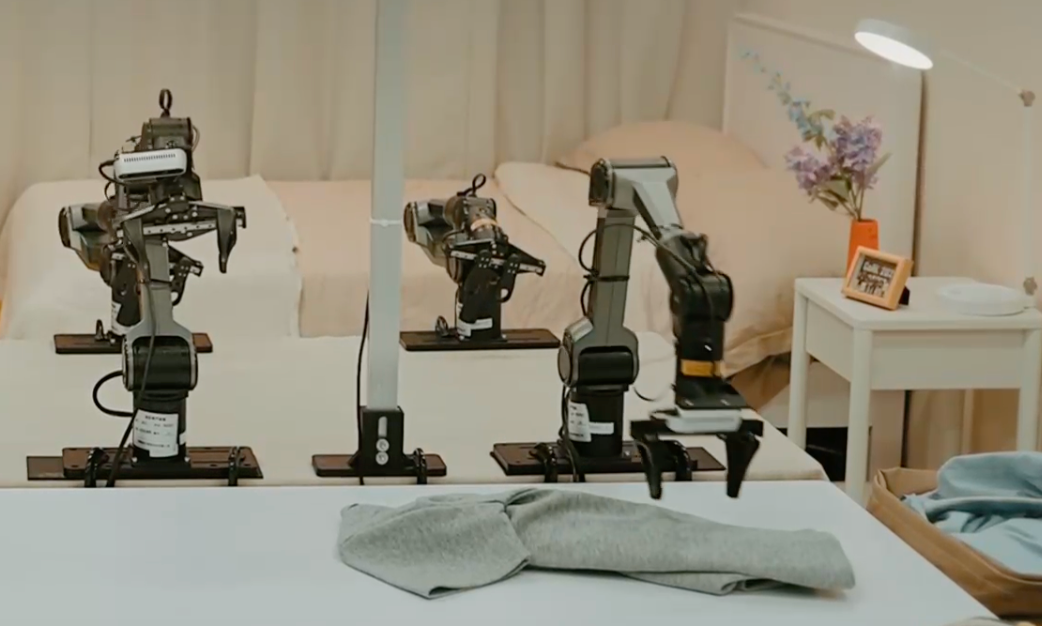} &
    \includegraphics[width=0.220\textwidth]{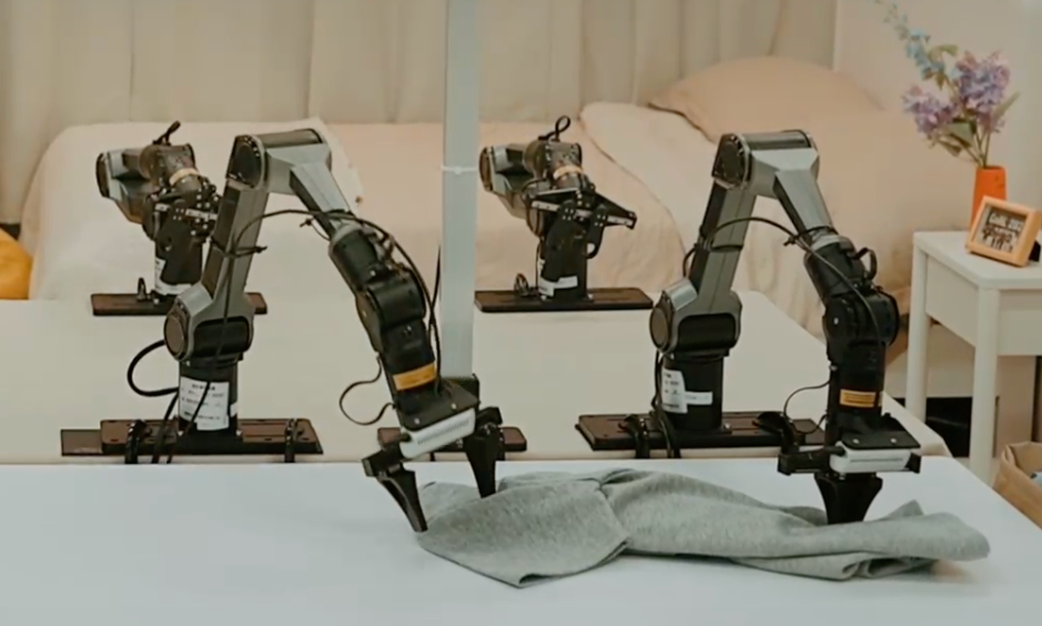} &
    \includegraphics[width=0.220\textwidth]{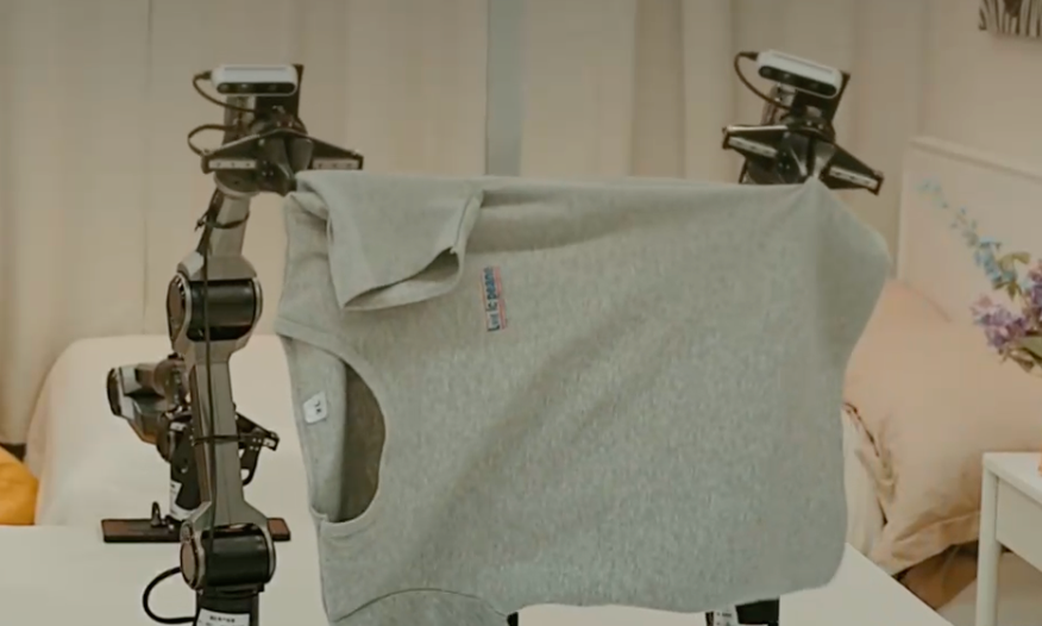} &
    \includegraphics[width=0.220\textwidth]{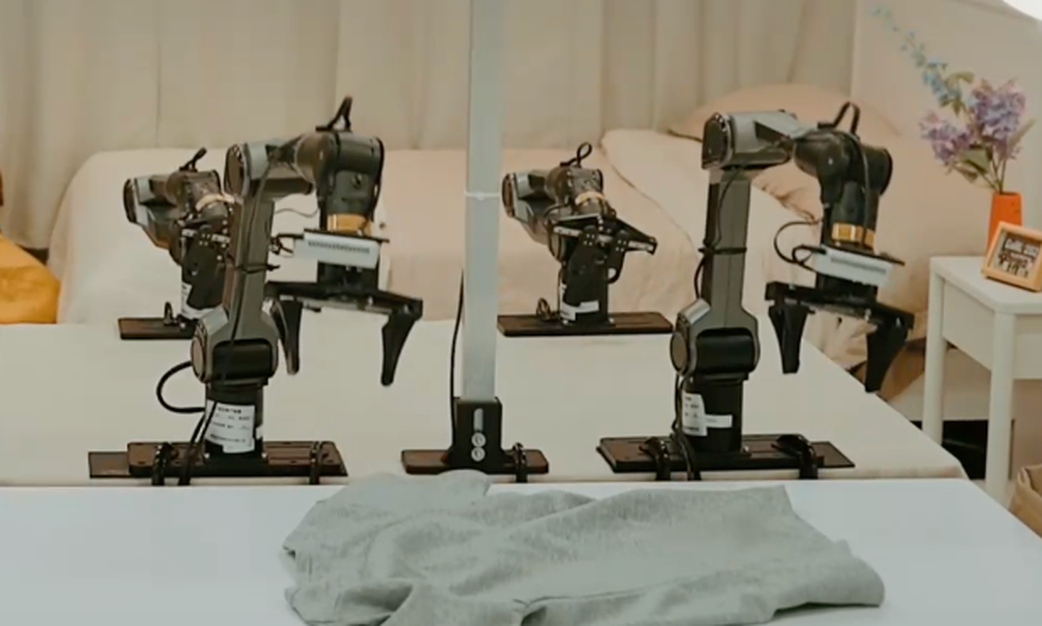} \\
    \makecell{Initial\\random wrinkles} & \makecell{Step 1\\grasp keypoints} & \makecell{Step 2\\flatten} & \makecell{Step 3\\lay down}
    \end{tabular}\par\medskip
    \begin{tabular}{ccc}
    \includegraphics[width=0.220\textwidth]{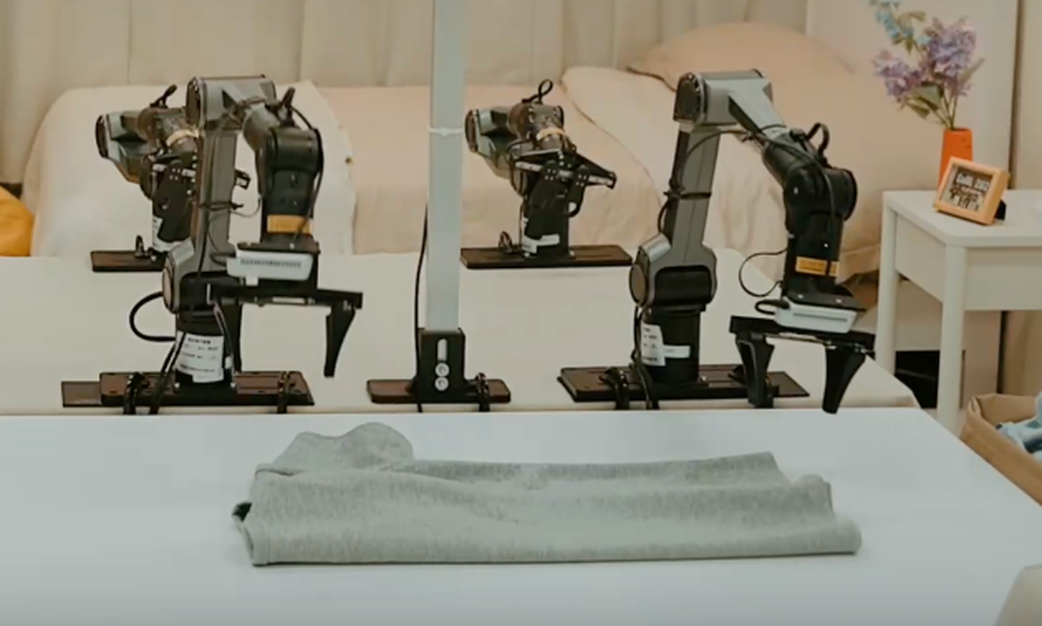} &
    \includegraphics[width=0.220\textwidth]{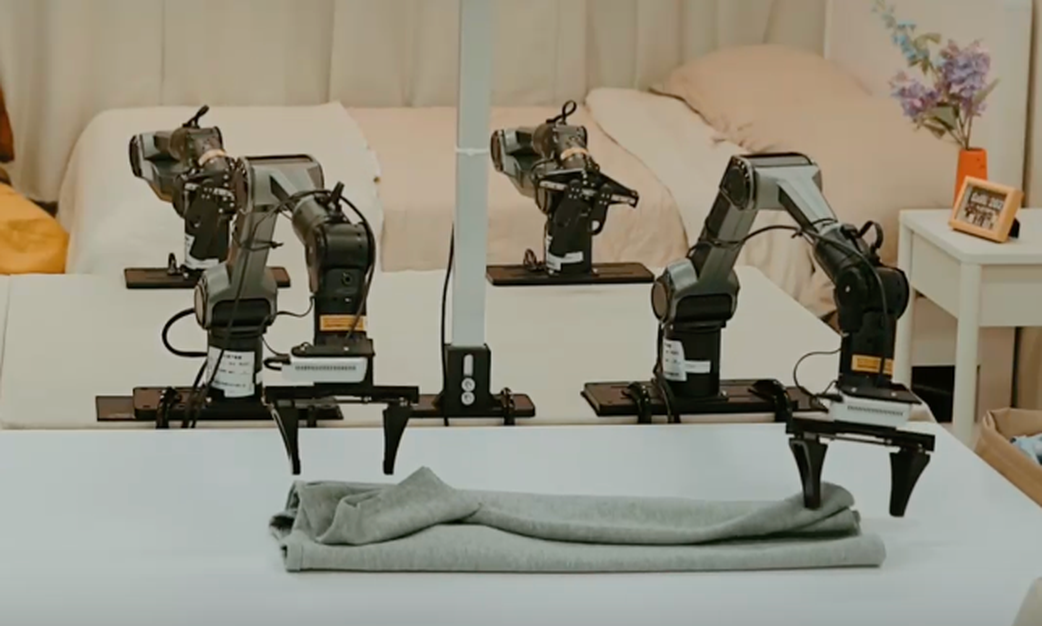} &
    \includegraphics[width=0.220\textwidth]{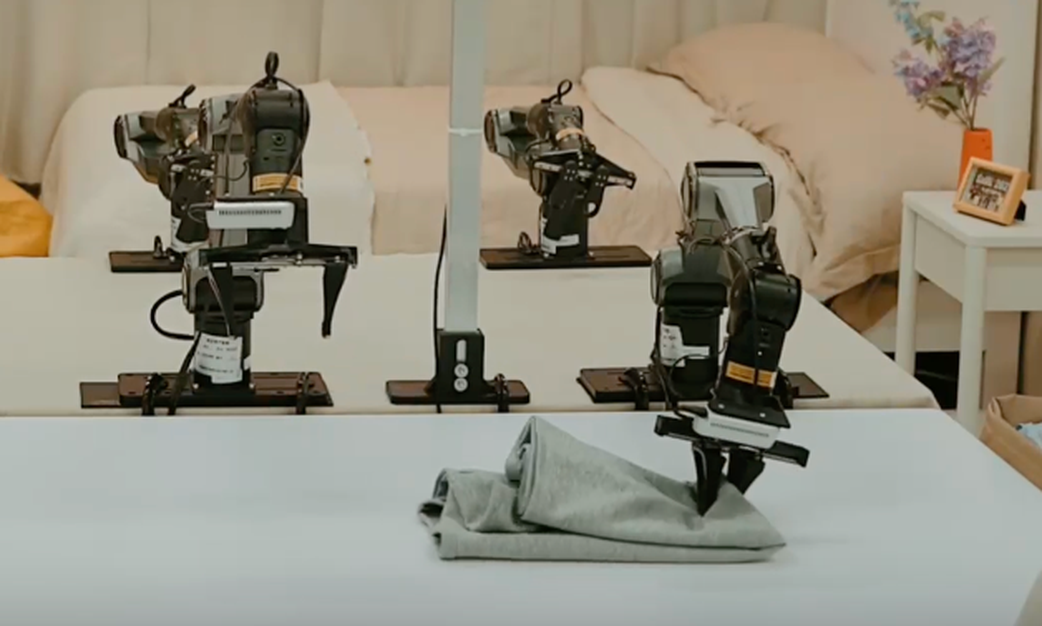} \\
    \makecell{Step 4\\first fold} & \makecell{Step 5\\second fold} & \makecell{Step 6\\third fold}
    \end{tabular}
    \caption{\textbf{Stage sequence of the bimanual garment-folding task.}}
    \label{fig:task_pipeline}
\end{figure}

Each trial starts from a randomly wrinkled garment.
The two arms successively grasp keypoints, flatten the garment, lay it down, and execute three folds.
\figref{fig:task_pipeline} shows the complete workflow, and \figref{fig:garment_conditions} shows the three garment conditions.
All methods use the same task definition and success criteria.

\FloatBarrier
\subsubsection{Evaluation Metrics}

Evaluation metrics cover task performance, action-chunk continuity, system latency, and physical tracking, measuring task completion, command continuity, temporal misalignment, and reference-trajectory tracking, respectively.
These metrics are not interchangeable.

\paragraph{Task-level metrics}
For $N$ trials of a method, let $S$ denote the number of successes and $T_i$ the duration of trial $i$.
Success rate and mean execution time are
\begin{equation}
    R_{\mathrm{succ}}=\frac{S}{N},\qquad
    \bar{T}=\frac{1}{N}\sum_{i=1}^{N}T_i .
    \label{eq:task_metrics}
\end{equation}
To account for both success probability and execution speed, we also report successful-task throughput,
\begin{equation}
    \mathcal{Q}=R_{\mathrm{succ}}\frac{3600}{\bar{T}},
    \label{eq:throughput}
\end{equation}
in successful tasks per hour.
We report a 95\% Wilson confidence interval for the success rate of each method across its 30 trials.
The experiments are not paired, so we do not conduct between-method significance tests.

\paragraph{Action-chunk continuity and smoothness metrics}
Let $a_k\in\mathbb{R}^{m}$ be the action published to the robot controller at time $k$, and let $\hat{a}^{\mathrm{tail}}_k$ be the time-aligned prediction from the previous chunk at the same time.
For the valid-overlap set $\Omega$, the instantaneous trajectory deviation is
\begin{equation}
    d_k=\frac{1}{m}\sum_{j=1}^{m}\left|a_{k,j}-\hat{a}^{\mathrm{tail}}_{k,j}\right| .
    \label{eq:instant_gap}
\end{equation}
Tail Gap averages the deviation over all valid-overlap times, whereas Switch Gap uses only the handover set $\mathcal{B}$:
\begin{equation}
    G_{\mathrm{tail}}=\frac{1}{|\Omega|}\sum_{k\in\Omega}d_k,\qquad
    G_{\mathrm{switch}}=\frac{1}{|\mathcal{B}|}\sum_{k\in\mathcal{B}}d_k .
    \label{eq:chunk_gaps}
\end{equation}
A low Switch Gap therefore indicates a well-connected handover instant but does not guarantee agreement over the subsequent predicted tail.

For a single joint, $m=1$, and the metrics reduce to mean absolute joint-position deviations in rad.
Dynamic command smoothness is evaluated using finite-difference velocity and acceleration at control period $h$:
\begin{equation}
    v_k=\frac{a_k-a_{k-1}}{h},\qquad
    \alpha_k=\frac{a_k-2a_{k-1}+a_{k-2}}{h^2}.
    \label{eq:finite_difference}
\end{equation}
We report maximum absolute acceleration in representative trials.
Tail Gap and Switch Gap characterize geometric continuation between chunks, whereas maximum acceleration measures dynamic impact at switching.
The quantities are not equivalent.

\subsubsection{Comparative Analysis}

As shown in \tabref{tab:task_results}, the six methods differ markedly on long-horizon bimanual garment folding.
Legato and VLASH achieve the strongest overall task performance, completing $29/30$ and $28/30$ trials, respectively.
Legato leads all three task metrics with $96.7\%$ success, $73.56$ s mean completion time, and $47.31~\mathrm{h}^{-1}$ successful-task throughput.
VLASH ranks second overall, with $93.3\%$ success and $43.22~\mathrm{h}^{-1}$ throughput.

Legato performs best among training-based methods, while Temporal Smoothing provides the strongest overall performance among training-free methods, with $76.7\%$ success and $29.38~\mathrm{h}^{-1}$ throughput.
Training-time RTC, Naive Asynchronous, and Inference-time RTC achieve success rates no higher than $63.3\%$ and mean completion times exceeding $120$ s.
Notably, Inference-time RTC does not outperform the other methods despite inference at $3$ Hz.
A higher replanning rate alone therefore does not guarantee better task performance; the outcome also depends on the interaction between chunk continuity, state mismatch, and execution mechanisms.

\begin{table}[!htbp]
\caption{\textbf{Task-level results for real-time VLA execution methods.}}
\label{tab:task_results}
\centering
\tablefont
\renewcommand{\arraystretch}{1.10}
\begin{tabular}{lccccccc}
\toprule
\textbf{Method} & \makecell{\textbf{Inference}\\\textbf{(Hz)}} & \makecell{\textbf{Publication}\\\textbf{(Hz)}} & \textbf{NFE} & \textbf{Success rate} & \makecell{\textbf{95\% Wilson}\\\textbf{interval}} & \makecell{\textbf{Mean time}\\\textbf{(s)}} & \makecell{\textbf{Throughput}\\\textbf{(h$^{-1}$)}} \\
\midrule
VLASH & 2 & 30 & 10 & 28/30 (93.3\%) & [78.7, 98.2] & 77.73 & 43.22 \\
Training-time RTC & 1 & 30 & 10 & 19/30 (63.3\%) & [45.5, 78.1] & 129.37 & 17.62 \\
Legato & 2 & 50 & 5 & \textbf{29/30 (96.7\%)} & [83.3, 99.4] & \textbf{73.56} & \textbf{47.31} \\
Temporal Smoothing & 3 & 30 & 10 & 23/30 (76.7\%) & [59.1, 88.2] & 93.93 & 29.38 \\
Naive Asynchronous & 1 & 30 & 10 & 19/30 (63.3\%) & [45.5, 78.1] & 124.13 & 18.37 \\
Inference-time RTC & 3 & 30 & 10 & 18/30 (60.0\%) & [42.3, 75.4] & 133.73 & 16.15 \\
\bottomrule
\end{tabular}

\end{table}

\FloatBarrier
\subsubsection{Action-Chunk Continuity}

We use the third left-arm joint, \texttt{left\_j3}, as a representative joint to analyze action continuity.
\tabref{tab:continuity} shows clear differences in action-continuity metrics among the real-time execution methods.
Legato achieves the lowest Tail Gap and Switch Gap, $0.0193$ rad and $0.0062$ rad, indicating reduced tail error and handover discontinuity.
VLASH also performs well, with gaps of $0.0459$ rad and $0.0173$ rad.
By contrast, Naive Asynchronous reaches a maximum acceleration of $744.1~\mathrm{rad/s^2}$ and has large values for both gaps, showing that direct asynchronous chunk replacement can cause pronounced trajectory discontinuities.

\begin{table}[!htbp]
\caption{\textbf{Action-continuity results for \texttt{left\_j3}; lower is better for every metric.}}
\label{tab:continuity}
\centering
\renewcommand{\arraystretch}{1.05}
\tablefont
\setlength{\tabcolsep}{8pt}
\begin{tabular}{@{}lccc@{}}
\toprule
\textbf{Method} & \makecell{\textbf{Maximum acceleration}\\\textbf{(rad/s$^2$)}} & \makecell{\textbf{Tail Gap}\\\textbf{(rad)}} & \makecell{\textbf{Switch Gap}\\\textbf{(rad)}} \\
\midrule
Naive Asynchronous & 744.1 & 0.1799 & 0.1538 \\
Temporal Smoothing & \textbf{25.6} & 0.0967 & 0.0998 \\
Inference-time RTC & 65.55 & 0.1310 & 0.1164 \\
Training-time RTC & 392.8 & 0.1201 & 0.0630 \\
Legato & 46.6 & \textbf{0.0193} & \textbf{0.0062} \\
VLASH & 73.9 & 0.0459 & 0.0173 \\
\bottomrule
\end{tabular}
\end{table}

Temporal Smoothing achieves the lowest maximum acceleration, $25.6~\mathrm{rad/s^2}$, reflecting strong local smoothing capability, but its Tail Gap and Switch Gap remain substantially higher than Legato's.
Maximum acceleration primarily reflects local smoothness, whereas Tail Gap and Switch Gap more directly characterize tail deviation and inter-chunk handover continuity.
A single smoothness metric is therefore insufficient to evaluate real-time VLA execution quality.

\FloatBarrier
\subsubsection{Local Action-Chunk Trajectories}

For each method, we select a 5 s window from a representative trial and plot position, velocity, and acceleration for \texttt{left\_j3}.
Colored solid curves indicate successive published action chunks, and black dotted curves show the time-aligned predicted tail of the previous chunk.
Gray shading and vertical dashed lines mark inference intervals and chunk handovers, respectively.

\begin{figure}[H]
    \centering
    \includegraphics[width=0.88\textwidth]{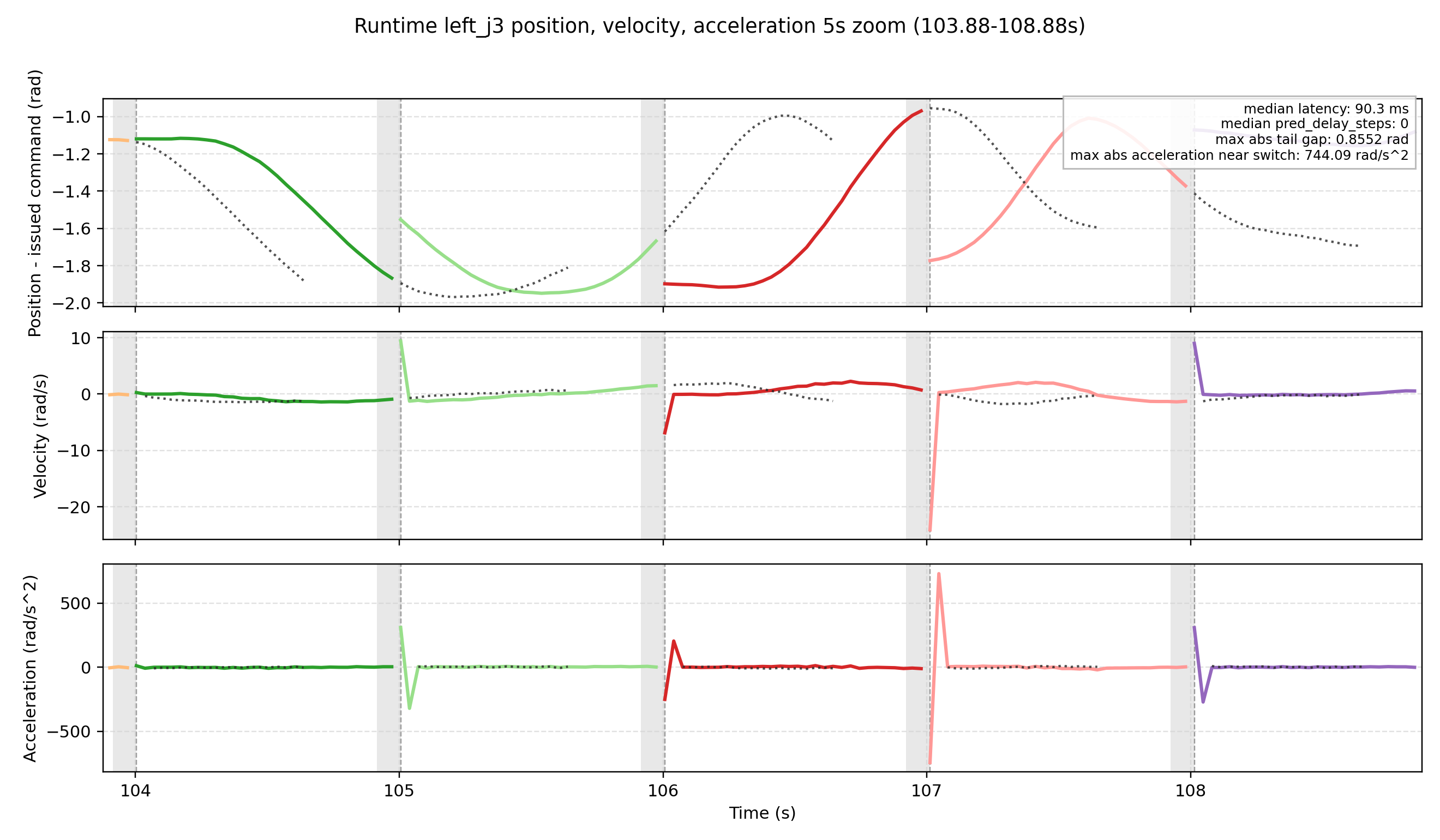}
    \caption{\textbf{Local \texttt{left\_j3} trajectory for Naive Asynchronous Execution.}}
    \label{fig:local_naive}
\end{figure}

\begin{figure}[H]
    \centering
    \includegraphics[width=0.88\textwidth]{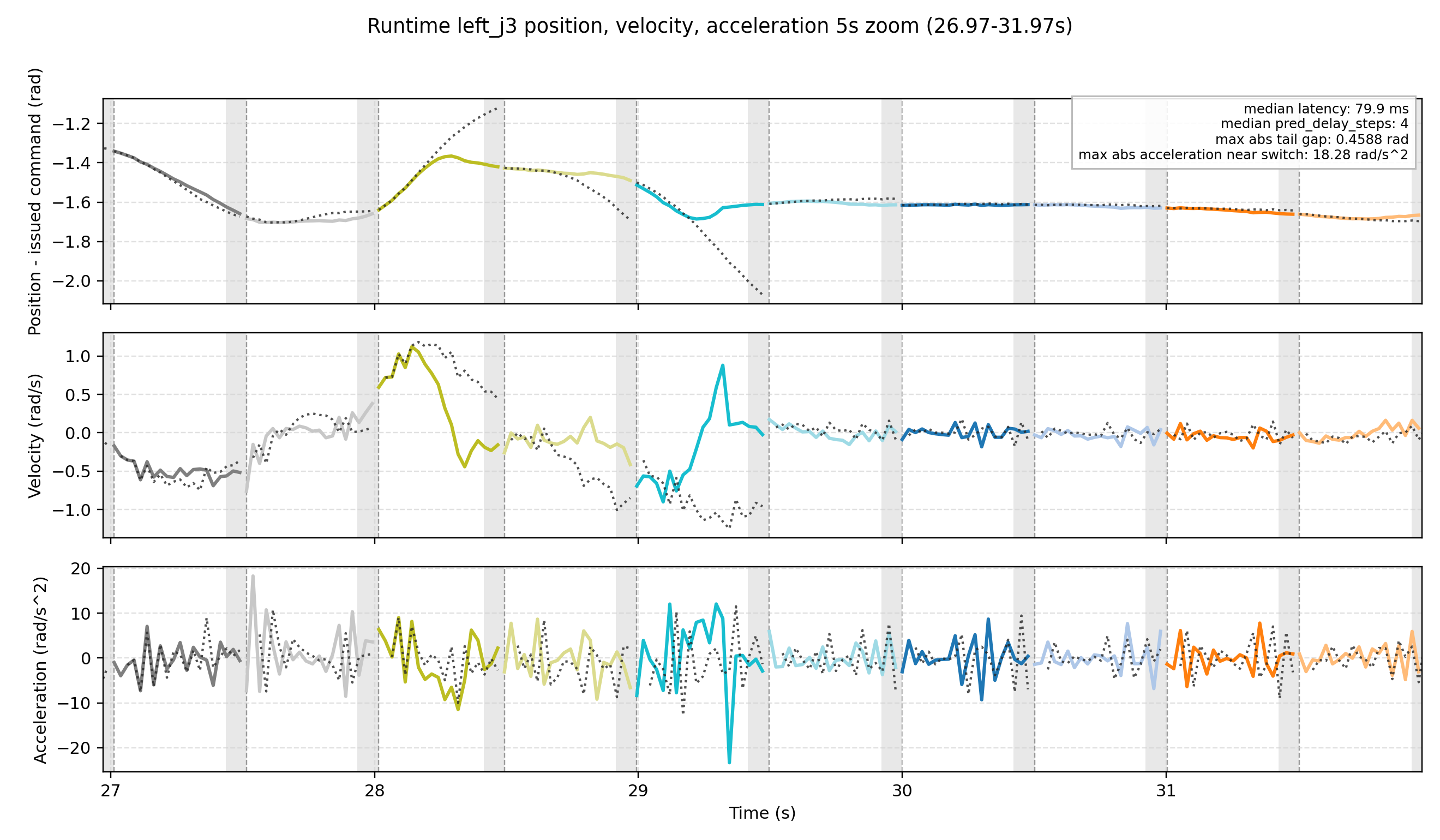}
    \caption{\textbf{Local \texttt{left\_j3} trajectory for Legato.}}
    \label{fig:local_legato}
\end{figure}

Figures~\ref{fig:local_naive}, \ref{fig:local_smoothing}, \ref{fig:local_inference_rtc}, \ref{fig:local_training_rtc}, \ref{fig:local_vlash}, and \ref{fig:local_legato} show the six local trajectories.
Near 107 s, Naive Asynchronous jumps from $-1.0$ rad to $-1.8$ rad, with velocity falling to approximately $-25$ rad/s.
Maximum acceleration and maximum absolute tail deviation reach 744.09 rad/s$^2$ and 0.8552 rad, respectively.
Temporal Smoothing reduces these values to 10.27 rad/s$^2$ and 0.6751 rad, primarily suppressing instantaneous impact.

Inference-time RTC exhibits a discontinuity near 13.7 s, with values of 65.55 rad/s$^2$ and 0.1813 rad.
Training-time RTC shows a velocity reset near 84--87 s, with values of 197.16 rad/s$^2$ and 0.8382 rad.
These events indicate that insufficient prefix or delay alignment can still produce handover impulses.

VLASH yields values of 34.58 rad/s$^2$ and 0.4615 rad.
Legato keeps velocity mostly within $[-1,1]$ rad/s, with values of 18.28 rad/s$^2$ and 0.4588 rad.
Overall, suppressing switching impulses and reducing inter-chunk trajectory deviation are related but distinct objectives.

\FloatBarrier
\subsection{Experimental Evaluation of Two-Stage 2-NFE Denoising}
\label{subsec:two_stage_experiments}

We evaluate two-stage non-uniform denoising in terms of inference efficiency, offline action error, and combinations with execution mechanisms.
Under fixed offline inputs, we compare inference time, joint-space action error, and bimanual Tool Center Point (TCP) error with standard 10-NFE Flow to quantify the computational gains and error changes from fewer sampling steps.
We then combine the strategy with Legato and Temporal Smoothing to assess model efficiency, task performance, and action continuity under different real-time execution mechanisms.

\subsubsection{Experimental Analysis}

As shown in \tabref{tab:two_stage_timing}, the two-stage strategy reduces NFE from 10 to 2 and inference time from $61.557$ ms to $21.956$ ms.
This corresponds to a $2.804\times$ speedup and a $64.33\%$ reduction in time.

\begin{table}[!htp]
\caption{\textbf{Inference time for standard Flow and two-stage denoising.}}
\label{tab:two_stage_timing}
\centering
\tablefont
\renewcommand{\arraystretch}{1.05}
\begin{tabular*}{0.92\textwidth}{@{\extracolsep{\fill}}lcccc@{}}
\toprule
\textbf{Sampler} & \textbf{NFE} & \textbf{Mean (ms)} & \textbf{Speedup} & \textbf{Time reduction} \\
\midrule
Standard Flow & 10 & 61.557 & $1.000\times$ & 0 \\
Two-stage non-uniform & 2 & \textbf{21.956} & \textbf{$2.804\times$} & \textbf{64.33\%} \\
\bottomrule
\end{tabular*}
\end{table}

\begin{table}[!htp]
\caption{\textbf{Action-error comparison.}}
\label{tab:two_stage_offline_accuracy}
\centering
\tablefont
\renewcommand{\arraystretch}{1.05}
\setlength{\tabcolsep}{5pt}
\begin{tabular}{clcccc}
\toprule
\textbf{Data} & \textbf{Model} & \makecell{\textbf{Joint MAE}\\\textbf{(rad)}} & \makecell{\textbf{Joint RMSE}\\\textbf{(rad)}} & \makecell{\textbf{Left TCP}\\\textbf{(mm)}} & \makecell{\textbf{Right TCP}\\\textbf{(mm)}} \\
\midrule
\multirow{2}{*}{\makecell[l]{Offline\\sequence I}} & Standard Flow & 0.008525 & \textbf{0.012877} & \textbf{11.396} & 7.676 \\
 & Two-stage & \textbf{0.008396} & 0.013329 & 11.493 & \textbf{6.461} \\
\midrule
\multirow{2}{*}{\makecell[l]{Offline\\sequence II}} & Standard Flow & \textbf{0.006503} & \textbf{0.008931} & 6.930 & \textbf{5.684} \\
 & Two-stage & 0.006797 & 0.009104 & \textbf{4.831} & 6.718 \\
\bottomrule
\end{tabular}
\end{table}

\tabref{tab:two_stage_offline_accuracy} summarizes joint error and bimanual TCP translation error on two fixed validation sequences.
The methods trade advantages across sequences and metrics, but their errors remain at the same order of magnitude.
Under the current fixed offline replay, two-stage 2-NFE sampling reduces model-inference time while retaining an action-error scale comparable to standard 10-NFE Flow.
Appendix~\ref{app:two_stage_fixed_sequence} provides detailed per-joint results.

\FloatBarrier
\subsubsection{Combination with Real-Time Execution Mechanisms}
\label{subsubsec:two_stage_runtime_compatibility}

We integrate two-stage 2-NFE denoising into Legato and Temporal Smoothing and compare it with their respective 5-NFE and 10-NFE baselines.
The experiments follow the task and evaluation protocol in Section~\ref{subsec:distributed_experiments}.

\begin{table}[!htp]
\caption{\textbf{Combined evaluation of two-stage 2-NFE denoising with Legato and Temporal Smoothing.} $\uparrow$ and $\downarrow$ indicate that higher and lower values are better, respectively.}
\label{tab:two_stage_runtime_compatibility}
\centering
\tablefont
\renewcommand{\arraystretch}{1.10}
\setlength{\tabcolsep}{3pt}
\begin{tabularx}{\linewidth}{@{}p{3.0cm}lccccY@{}}
\toprule
& & \multicolumn{3}{c}{\textbf{Model efficiency}} & \multicolumn{2}{c}{\textbf{Task performance}} \\
\cmidrule(lr){3-5}\cmidrule(lr){6-7}
\textbf{Method} & \textbf{Config.} & \textbf{NFE} &
\makecell{\textbf{Inference time}\\\textbf{(ms) $\downarrow$}} &
\makecell{\textbf{Relative}\\\textbf{speedup $\uparrow$}} &
\makecell{\textbf{Success rate}\\\textbf{(\%) $\uparrow$}} &
\makecell{\textbf{Successful-trial}\\\textbf{time (s) $\downarrow$}} \\
\midrule
Legato & Baseline & 5 & $36.808$ & $1.000\times$ & 29/30 (96.7) & 73.56 \\
\makecell[l]{Legato\\+ two-stage 2-NFE} & Combined & 2 & $21.956$ & $1.676\times$ & 26/30 (86.7) & 76.13 \\
\midrule
Temporal Smoothing & Baseline & 10 & $61.557$ & $1.000\times$ & 23/30 (76.7) & 93.93 \\
\makecell[l]{Temporal Smoothing\\+ two-stage 2-NFE} & Combined & 2 & $21.956$ & $2.804\times$ & 21/30 (70.0) & 98.63 \\
\bottomrule
\end{tabularx}
\par\vspace{4pt}
\begin{tabularx}{\linewidth}{@{}p{3.0cm}lYYY@{}}
\toprule
& & \multicolumn{3}{c}{\textbf{Execution continuity}} \\
\cmidrule(lr){3-5}
\textbf{Method} & \textbf{Config.} &
\makecell{\textbf{Tail Gap}\\\textbf{(rad) $\downarrow$}} &
\makecell{\textbf{Switch Gap}\\\textbf{(rad) $\downarrow$}} &
\makecell{\textbf{Representative-trial}\\\textbf{max. $|\alpha|$ (rad/s$^2$) $\downarrow$}} \\
\midrule
Legato & Baseline & 0.0193 & 0.0062 & 46.6 \\
\makecell[l]{Legato\\+ two-stage 2-NFE} & Combined & 0.0232 & 0.0074 & 55.9 \\
\midrule
Temporal Smoothing & Baseline & 0.0967 & 0.0998 & 25.6 \\
\makecell[l]{Temporal Smoothing\\+ two-stage 2-NFE} & Combined & 0.1160 & 0.1198 & 30.7 \\
\bottomrule
\end{tabularx}
\end{table}

As shown in \tabref{tab:two_stage_runtime_compatibility}, both combined configurations reduce inference time to $21.956$ ms, yielding speedups of $1.676\times$ and $2.804\times$, respectively.
Relative to their baselines, success rates fall from $96.7\%$ to $86.7\%$ and from $76.7\%$ to $70.0\%$.
Mean successful-trial time increases by approximately $5\%$ in both cases, and all three continuity metrics increase by approximately $20\%$.
Thus, two-stage 2-NFE denoising can be directly combined with different real-time execution mechanisms and substantially reduce model-side action-inference cost.
However, more aggressive NFE compression still incurs some closed-loop performance loss, reflecting a trade-off between inference efficiency and action-generation quality.

%% file: sections/02_related_work_en.tex
\section{Literature Review}
\label{sec:related_work}

As VLA models become widely adopted in robotics, the mismatch between low-rate inference and high-rate continuous control becomes a central real-time bottleneck. With action chunking, each model call typically predicts a future action sequence, while visual acquisition, state feedback, network transmission, model inference, and robot execution operate at different time scales. Real-time VLA research has therefore expanded beyond reducing inference time toward joint optimization of perception, inference, action generation, chunk transition, replanning, and robot execution.

\begin{figure}[!htbp]
    \centering
    \includegraphics[width=\textwidth,trim=0 40 0 60,clip]{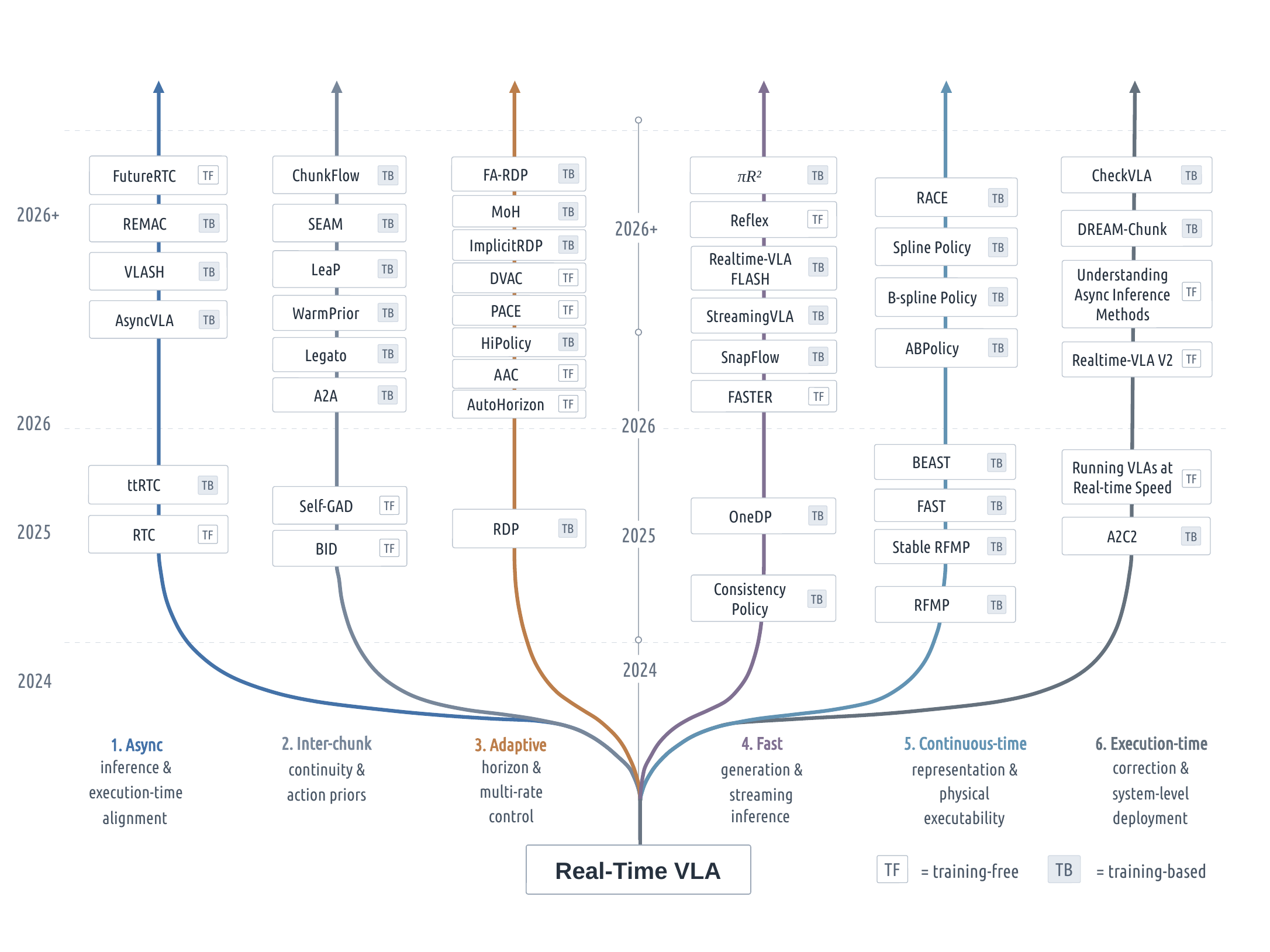}
    \caption{\textbf{Chronological development and methodological taxonomy of representative real-time VLA studies.} The six branches denote asynchronous inference and execution-time alignment, inter-chunk continuity and action priors, adaptive execution horizons and multi-rate control, fast action generation and streaming inference, continuous-time action representations and physical executability, and execution-time correction and system-level deployment. Only representative studies are shown; publication dates and version information follow the cited references.}
    \label{fig:realtime_vla_evolution}
\end{figure}

Figure~\ref{fig:realtime_vla_evolution} organizes this research by its principal technical problems. We divide the literature into six categories: (1) asynchronous inference and execution-time alignment; (2) inter-chunk continuity and action priors; (3) adaptive execution horizons and multi-rate control; (4) fast action generation and streaming inference; (5) continuous-time action representations and physical executability; and (6) execution-time correction and system-level deployment. Each study is assigned according to its primary contribution, although several methods address more than one category.

\subsection{Asynchronous Inference and Execution-Time Alignment}
\label{subsec:async_alignment}

Synchronous inference must wait for a new action chunk and can cause control pauses.
Asynchronous execution overlaps inference with motion but separates the observation time from the new chunk's handover time~\cite{rtc,vlash}.
Existing methods differ mainly in what they align.

Inference-time RTC and Training-time RTC treat actions that remain active during inference as a committed prefix, establishing consistent time indices between chunks.
RTC fixes this prefix during Flow Matching or diffusion sampling and generates only the actions after handover, reducing conflicts between independently generated chunks~\cite{rtc}.
Training-time RTC simulates random inference delays and committed-prefix lengths during training to learn continuation from a given prefix~\cite{ttrtc}.
These methods introduce prefix constraints at inference and training time, respectively, but neither explicitly compensates for robot-state changes during inference.

Aligned action indices do not guarantee consistent conditioning states at handover.
VLASH and FutureRTC therefore predict the conditions at action handover to bring model inputs closer to the future execution context.
VLASH propagates proprioceptive state to the expected handover time using the actions still being executed, then conditions the next VLA prediction on that future state~\cite{vlash}.
FutureRTC predicts both state and visual features at handover, extending alignment from proprioception to multimodal conditions~\cite{futurertc}.

REMAC and AsyncVLA extend alignment inside the action chunk, distinguishing known states or generation progress at different positions instead of treating the chunk as an independently and synchronously generated unit.
REMAC uses masked action chunking to correct and continue trajectories with partially known, missing, or perturbed actions, addressing tracking discrepancies between plans and execution~\cite{remac}.
AsyncVLA modifies the Flow Matching schedule so that action tokens occupy different generation stages and prioritizes low-confidence positions~\cite{asyncvla}.
From prefix constraints and future-condition prediction to differentiated within-chunk generation, temporal alignment thus extends beyond the handover boundary to the asynchronous generation process.

\subsection{Inter-Chunk Continuity and Action Priors}
\label{subsec:continuity_prior}

Consecutive diffusion or Flow Matching policy calls~\cite{diffusionpolicy} usually start from independent random variables. Even with nearly unchanged observations, adjacent chunks can converge to different feasible modes, causing discontinuities in position, velocity, acceleration, or higher-order dynamics. The research question has evolved from smoothing a boundary to converting action history into an effective prior for the next generation process.

BID\cite{bid} uses guided test-time sampling to generate several candidate chunks. Backward consistency and forward-quality criteria select a trajectory that continues the previous decision without sacrificing future planning quality for local smoothness. Self-GAD\cite{selfgad} changes post-sampling selection into guidance during sampling. It constructs a prior from the previous decision and steers denoising toward temporally consistent solutions with limited extra sampling cost.

Subsequent methods incorporate continuity directly into learning or inference. Legato\cite{legato} initializes training consistently with the deployed action-continuation plan, teaching a Flow policy to extend existing trajectories. SEAM\cite{seam} uses the unexecuted tail of the previous chunk as a motion reference and analytically corrects the current Flow velocity. ChunkFlow\cite{chunkflow} partitions actions into frozen, editable-overlap, and future regions, with explicit seam, velocity, and acceleration constraints during training. These methods move continuity modeling from positional stitching toward the generation process and higher-order kinematics.

Recent work redesigns the source distribution of generative policies. A2A\cite{a2a} replaces independent Gaussian noise with a representation of action history and learns transport from executed to future actions. WarmPrior\cite{warmprior} shortens the Flow transport path through a temporal action prior, bringing generation into closer agreement with correlations between adjacent time points. LeaP\cite{leap} learns a state-conditioned source prior whose mean and uncertainty adapt to the robot state. Inter-chunk research is therefore shifting from smoothing generated outputs to selecting an appropriate starting point for generation. An overly strong historical prior can nevertheless propagate and amplify errors after a faulty chunk or abrupt environmental change. Online assessment of history reliability and prior strength remains necessary.

\subsection{Adaptive Execution Horizons and Multi-Rate Control}
\label{subsec:adaptive_horizon}

This line of work adapts the execution horizon using system state, action structure, or model uncertainty to balance frequent inference at short horizons against action drift at long horizons.

AAC\cite{aac} adjusts the execution length from predictive uncertainty, using longer prefixes when confidence is high and requesting new observations when uncertainty increases. Without relying on internal model features, PACE\cite{pace} detects deceleration, pauses, and phase transitions in the action velocity profile as natural replanning boundaries. DVAC\cite{dvac} extracts stability from Flow Matching denoising by comparing the variance of late clean-action estimates and executing only the stable, low-variance prefix. AutoHorizon\cite{autohorizon} estimates a prediction's valid boundary from self-attention among VLA action tokens. These approaches answer the replanning question through action uncertainty, motion phase, generation stability, and internal attention, respectively.

Beyond adapting one horizon, MoH\cite{moh} jointly models several action horizons and executes near-term actions according to cross-scale consensus. HiPolicy\cite{hipolicy} predicts at multiple control rates. A low-rate branch provides coarse long-horizon plans, while a high-rate branch produces precise actions. Action-chunk length, execution horizon, and control rate are thus treated as coupled deployment time-scale parameters rather than fixed policy hyperparameters.

Contact-rich tasks further motivate slow--fast coordination and multimodal, multi-rate control. RDP\cite{rdp} combines a low-rate visual diffusion policy with high-rate tactile or force-feedback control, preserving long-horizon visual planning capability while responding quickly to changes in contact. ImplicitRDP\cite{implicitrdp} models low-rate vision and high-rate force feedback end to end using a unified network and causal attention. FA-RDP\cite{fardp} changes generation rate and sampling steps according to task phase and action multimodality. It preserves multimodal planning in free space and increases closed-loop updates during contact. These methods move real-time VLA from a fixed execution cycle toward a phase-, state-, and modality-aware adaptive control clock.

\subsection{Fast Action Generation and Streaming Inference}
\label{subsec:fast_streaming}

Diffusion and Flow policies typically require repeated network evaluations and numerical integration to generate a complete action chunk. This line of work therefore follows two routes: reducing generation steps and changing output timing.

Consistency Policy\cite{consistency} uses consistency distillation to compress a multi-step diffusion policy into one or a few steps, enabling high-rate generative robot policies. OneDP\cite{onedp} applies one-step distribution distillation, generating a complete action sequence in a single forward pass. For Flow Matching policies, SnapFlow\cite{snapflow} uses progressive self-distillation to learn mappings across longer Flow-time intervals. It compresses iterative ODE integration into one or a few network calls. These methods primarily reduce the cost of generating a complete action chunk.

Total chunk-generation time does not directly determine closed-loop responsiveness. FASTER\cite{faster} therefore emphasizes Time to First Action. Horizon-aware scheduling completes and releases near-term actions first while refining later actions in the background. $\pi R^2$\cite{pir2} decouples slow visual--language conditioning from fast proprioceptive conditioning. Positions within its action buffer can remain at different generation stages, continuously exposing stabilized near-term actions. These methods shift the goal from producing a complete chunk faster to obtaining the next useful action earlier.

StreamingVLA\cite{streamingvla} further relaxes the serial observe--generate--execute loop by overlapping observation, Flow generation, and execution. Reflex\cite{reflex} improves sustained inference efficiency through caching, asynchronous vision--action pipelines, and optimized operators. Realtime-VLA FLASH\cite{realtimeflash} adopts speculative inference. A lightweight draft path proposes actions quickly, while the main model verifies them and performs full inference only when needed. The resulting progression spans multi-step, one- or few-step, near-term-first, streaming, and speculative generation.

\subsection{Continuous-Time Action Representations and Physical Executability}
\label{subsec:continuous_physical}

The fifth category reconsiders real-time performance through action representation and physical robot capability. Conventional chunks are discrete action sequences defined at a fixed sampling rate. Higher control rates require longer and higher-dimensional outputs, yet discrete points do not guarantee continuous velocity, acceleration, or curvature. Frequency-domain, spline, manifold, and dynamic structures provide more compact and physically executable representations.

FAST\cite{fast} compresses high-frequency action sequences into the frequency domain using the discrete cosine transform. Fewer action tokens can represent a complete temporal sequence, shortening autoregressive VLA outputs. BEAST\cite{beast} encodes actions with B-spline parameters, replacing per-step tokens with a fixed number of spline parameters. Continuous reconstruction then yields smoother high-frequency control. Both methods reduce output burden through high-frequency action compression and structured tokenization.

ABPolicy\cite{abpolicy} performs Flow Matching directly in B-spline control-point space. Asynchronous inference and spline refitting connect consecutive predictions continuously. B-spline Policy\cite{bspline} defines outputs as continuous-time splines, allowing one trajectory to be sampled at different control rates and rescaled in time. Spline Policy\cite{spline} treats a spline as a queryable, editable, and constrainable trajectory object. Compared with discrete chunks, these methods partly decouple control frequency from a fixed output grid and expose interfaces for velocity, acceleration, and local trajectory correction.

Another route incorporates robot geometry or dynamics. RFMP\cite{rfmp} defines Flow Matching on the Riemannian manifold of robot states, aligning probability paths, velocity fields, and integration with non-Euclidean action geometry. Stable RFMP\cite{stablerfmp} adds stability constraints to improve preservation of target motion distributions under perturbation. RACE\cite{race} uses time-optimal action-chunk execution based on robot reachability and execution speed, exceeding demonstration speed while satisfying motion constraints. Real-time VLA must therefore consider both action-generation time and whether the robot can execute the generated actions at higher rates and speeds.

\subsection{Execution-Time Correction and System-Level Deployment}
\label{subsec:execution_system}

The sixth category addresses online correction, execution verification, and end-to-end real-time performance after generation or execution has begun. A valid chunk can become unsuitable because of tracking error, target motion, contact slip, or new visual evidence. Optimizing only generation latency and continuity cannot ensure closed-loop robustness on a physical system.

A2C2\cite{a2c2} inserts a lightweight correction head between a low-rate base VLA and high-rate control. The head reads the latest observation every control cycle and applies residual corrections to the current chunk, restoring local closed-loop feedback without repeatedly invoking the full VLA. CheckVLA\cite{checkvla} formulates the problem as execution-time verification. An action-conditioned world model predicts expected observation changes during execution and triggers suffix repair when actual and predicted observations differ substantially. DREAM-Chunk\cite{dreamchunk} uses a latent world model to evaluate future states for several candidate chunks and selects candidates consistent with the evolving environment using actual rollout feedback.

Real-time VLA has also expanded from an algorithmic problem into a complete deployment-system problem. Running VLAs at Real-time Speed\cite{runningrt} jointly optimizes model computation, caching, data transfer, and streaming execution. It shows that high-rate visual processing and continuous action output require coordinated model and runtime design. Realtime-VLA V2\cite{realtimev2} covers camera and proprioceptive calibration, action buffering, trajectory interpolation, inference services, and robot control. It emphasizes that sensor, communication, and execution latency influence physical performance alongside model latency. Understanding Asynchronous Inference Methods for Vision-Language-Action Models\cite{understandingasync} compares several asynchronous execution schemes under common base-model, task, and latency settings. It notes that the advantages of these methods depend strongly on actual latency, execution horizon, and control settings.

Real-time VLA methods now address asynchronous execution and chunk continuity, adaptive execution horizons, faster generation, continuous-time action representations, and execution-time correction.
Although they share a real-time objective, they modify different system components and are not readily comparable under one experimental setting.
Adaptive-horizon methods change execution length or replanning frequency; faster generation often changes sampling or training; continuous-time methods change the action representation; and execution-time correction may introduce additional feedback or prediction modules.
Combining all these approaches in one comparison would vary model structure, action representation, training, and execution configuration simultaneously, obscuring which mechanisms cause performance differences.

Our unified physical evaluation therefore focuses on two direct problems in asynchronous action-chunk execution: inference--execution misalignment and inter-chunk continuity.
We compare Naive Asynchronous, Temporal Smoothing, Inference-time RTC, Training-time RTC, Legato, and VLASH, all deployable with the same base policy, action representation, and robot interface.
Their shared model and execution pipeline allow a more focused analysis of how real-time execution mechanisms affect task performance and action continuity.

%% file: sections/07_conclusion_en.tex
\section{Conclusion}
\label{sec:conclusion}

We analyze latency at the model-computation and execution-chain levels to address the time-scale mismatch between low-rate VLA inference and high-rate continuous robot control.
The results show that real-time VLA performance is jointly constrained by model computation and multiple sources of system latency.

For the model-side bottleneck, we analyze velocity-field changes across Flow Matching integration stages.
The early and intermediate velocity fields remain broadly stable, whereas terminal integration exhibits stronger directional correction.
The resulting two-stage non-uniform sampling strategy uses the schedule $1\rightarrow0.3\rightarrow0$, reducing NFE from $10$ to $2$ and inference time from $61.557$ ms to $21.956$ ms.
This corresponds to a $2.804\times$ speedup and a $64.33\%$ reduction in time.

For system-side latency and asynchronous execution, we develop a distributed, thread-decoupled real-time inference and execution framework that jointly manages observation acquisition, model inference, action buffering, and runtime records.
Policy inference, action publication, and robot control rates can be configured independently.
Six representative real-time execution methods are deployed and evaluated on physical robots through this framework, demonstrating support for different execution mechanisms.

Under a common $\pi_{0.5}$ base policy, training set, and bimanual platform, long-horizon garment-folding experiments show that Legato and Temporal Smoothing provide the strongest overall performance among training-based and training-free methods, respectively.
Combining two-stage non-uniform sampling with these mechanisms yields inference speedups of $1.676\times$ and $2.804\times$, respectively.

\begingroup
\looseness=-1
Although this study focuses on VLA, the proposed non-uniform Flow sampling approach can also be extended to Flow Matching-based world--action models.
Future work will further investigate joint optimization of inference and execution scheduling through adaptive adjustment of inference rate, sampling budget, and action-handover strategies.
\par
\endgroup

%% file: sections/appendices_en.tex
\input{sections/appendix_a_en}
\input{sections/appendix_b_en}
\input{sections/appendix_c_en}
\input{sections/appendix_d_en}

%% file: sections/appendix_a_en.tex
\section{Task Performance Across Garment Conditions}
\label{app:per_material_results}

This appendix partitions the 180 physical trials in the main text by three garment conditions.
For each condition, it reports successful trials, mean completion time, and successful-task throughput.
The analysis complements the aggregate results by showing how each execution method behaves across manipulation objects.

\begin{table}[!htbp]
    \centering
    \tablefont
    \caption{\textbf{Task-level results across garment conditions.} Each row contains ten physical trials, and throughput is the number of successful tasks completed per hour.}
    \label{tab:per_material_results}
    \renewcommand{\arraystretch}{1.15}
    \setlength{\tabcolsep}{9pt}
    \begin{tabular}{llccc}
        \toprule
        \textbf{Method}
        & \makecell{\textbf{Garment}\\\textbf{condition}}
        & \textbf{Successes}
        & \makecell{\textbf{Mean time}\\\textbf{(s)}}
        & \makecell{\textbf{Throughput}\\\textbf{(h$^{-1}$)}} \\
        \midrule

        \multirow{3}{*}{VLASH}
        & Small purple & 10/10 & 56.4  & 63.83 \\
        & Large pink & 9/10  & 99.2  & 32.66 \\
        & Medium red & 9/10  & 77.6  & 41.75 \\
        \midrule

        \multirow{3}{*}{Training-time RTC}
        & Small purple & 8/10 & 116.5 & 24.72 \\
        & Large pink & 7/10 & 124.4 & 20.26 \\
        & Medium red & 4/10 & 147.2 & 9.78 \\
        \midrule

        \multirow{3}{*}{Legato}
        & Small purple & 10/10 & 72.47 & 49.68 \\
        & Large pink & 10/10 & 69.2  & 52.02 \\
        & Medium red & 9/10  & 79.0  & 41.01 \\
        \midrule

        \multirow{3}{*}{Temporal Smoothing}
        & Small purple & 8/10 & 84.1  & 34.24 \\
        & Large pink & 9/10 & 103.2 & 31.40 \\
        & Medium red & 6/10 & 94.5  & 22.86 \\
        \midrule

        \multirow{3}{*}{Naive Asynchronous}
        & Small purple & 6/10 & 134.4 & 16.07 \\
        & Large pink & 5/10 & 144.5 & 12.46 \\
        & Medium red & 8/10 & 93.5  & 30.80 \\
        \midrule

        \multirow{3}{*}{Inference-time RTC}
        & Small purple & 7/10 & 149.5 & 16.86 \\
        & Large pink & 7/10 & 129.5 & 19.46 \\
        & Medium red & 4/10 & 122.2 & 11.78 \\
        \bottomrule
    \end{tabular}
\end{table}

\tabref{tab:per_material_results} shows that Legato and VLASH complete 9/10--10/10 tasks under all three garment conditions, leading overall in success rate and throughput.
Legato maintains throughput of at least 41.01~h$^{-1}$, while VLASH achieves the highest value in the table, 63.83~h$^{-1}$, on the small purple garment.
Temporal Smoothing completes 6/10--9/10 tasks with throughput of 22.86--34.24~h$^{-1}$, performing well among training-free methods.
Both RTC methods complete only 4/10 tasks on the medium red garment.
These results indicate differences in sensitivity to garment size and appearance.

%% file: sections/appendix_b_en.tex
\section{Training Configuration}
\label{app:training_configuration}

\tabref{tab:training_configuration} summarizes the key settings needed to reproduce model training.
Only settings that directly affect training and reproducibility are retained; platform-management parameters such as logging frequency, checkpoint intervals, and runtime paths are omitted.

\begin{table}[!htbp]
    \centering
    \tablefont
    \caption{\textbf{Training configuration for two-stage non-uniform denoising.}}
    \label{tab:training_configuration}
    \renewcommand{\arraystretch}{1.18}
    \setlength{\tabcolsep}{10pt}
    \begin{tabularx}{\linewidth}{>{\raggedright\arraybackslash}p{0.27\textwidth} X}
        \toprule
        \textbf{Hyperparameter} & \textbf{Value} \\
        \midrule
        Training dataset & $3{,}048$ teleoperated demonstration trajectories \\
        Training steps & $30{,}000$ \\
        Batch size & $16$ \\
        Optimizer & AdamW, $\beta_1=0.9$, $\beta_2=0.95$, $\epsilon=10^{-8}$ \\
        Learning-rate schedule & Peak learning rate $2.5\times10^{-5}$; linear warmup for $1{,}000$ steps, followed by cosine decay to $2.5\times10^{-6}$ \\
        Regularization and gradient clipping & Weight decay $0.01$; maximum gradient norm $1.0$ \\
        \bottomrule
    \end{tabularx}
\end{table}

%% file: sections/appendix_c_en.tex
\section{Scale Consistency of the Flow Matching Velocity Field}
\label{app:single_episode_velocity}

Figure~\ref{fig:flow_stage_motivation} shows stage-dependent differences in velocity magnitude and directional correction during standard 10-NFE inference.
To assess whether predicted velocities match the scale of the training targets, this appendix uses the component-wise RMS defined in Eq.~\eqref{eq:flow_velocity_rms} to compare predicted and Flow Matching ground-truth (GT) velocities.
This complements the main-text analysis from the perspective of supervision scale.

\subsection{Scale Comparison with Flow Matching GT Velocity}

\begin{figure}[H]
    \centering
    \includegraphics[width=0.98\linewidth]{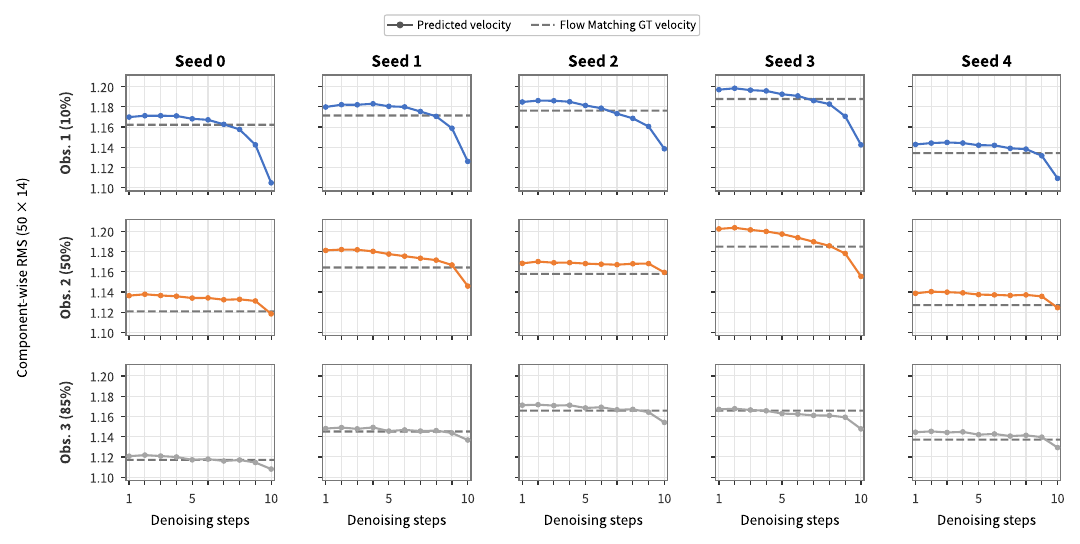}
    \caption{\textbf{Component-wise RMS comparison between predicted and Flow Matching GT velocity vectors in a representative validation sequence.} Colored solid curves show predicted velocity RMS; black horizontal dashed lines show the corresponding GT velocity RMS.}
    \label{fig:app_single_episode_teacher}
\end{figure}

We use the local velocity target $v^*$ defined in Section~\ref{sec:two_stage_denoising} as the Flow Matching GT velocity and compute its component-wise RMS in the valid 14-dimensional space.
Five fixed noise seeds at each of three progress positions ($0.10$, $0.50$, and $0.85$) in the same representative validation sequence yield 15 inference trajectories.
For a fixed action--noise pair, $v^*$ does not change across denoising steps, so its RMS appears as a horizontal dashed line in \figref{fig:app_single_episode_teacher}.
This velocity is a sample-level supervision target, not the magnitude of the dataset action itself.

Across 15 trajectories and 10 denoising steps, the mean absolute relative deviation is $0.775\%$, indicating comparable predicted and GT velocity RMS throughout integration.
At step 10, 14 trajectories have predicted velocity RMS slightly below the corresponding GT value.
The mean signed relative deviation is $-1.813\%$, with individual values ranging from $-4.941\%$ to $0.127\%$.
These results indicate that terminal updates mainly involve small velocity adjustments rather than reconstruction of the overall velocity magnitude.
This agrees with the broadly stable early and intermediate magnitudes and stronger terminal directional corrections in \figref{fig:flow_stage_motivation}.
The supervision-scale comparison complements the stage analysis and supports long-interval coarse generation followed by short-interval terminal refinement.

%% file: sections/appendix_d_en.tex
\section{Offline Action Accuracy of Two-Stage Non-Uniform Denoising}
\label{app:two_stage_fixed_sequence}

This appendix supplements offline sequence II in the main text with frame-wise joint trajectories, error distributions, and bimanual TCP pose errors.

\subsection{Comparison and Evaluation Protocol}

We use the model comparison and input protocol in Section~\ref{subsec:two_stage_experiments}.
The following details concern only frame-wise processing and metric computation for offline sequence II.

The fixed validation sequence contains 376 frames over 12.5 s.
Each frame independently uses its recorded observation.
Following the 14-dimensional action convention in the main text, the model's $50\times32$ output is aligned with the recorded action as \texttt{action\_chunk[0,:14]} versus \texttt{action[t,:14]}.
Errors for the 12 rotational joints are reported in rad, while the two grippers are reported separately in dataset command units.
Predicted state is not recursively passed to the next frame.

Bimanual TCP poses are computed by forward kinematics using the same robot Unified Robot Description Format (URDF).
For each arm, TCP position is the midpoint between the origins of the two fingertip links; orientation uses the corresponding sixth-axis end-link frame.
For frame $t$ and arm $a\in\{\mathrm{L},\mathrm{R}\}$, let $(\hat{\mathbf{p}}_{t,a},\hat{\mathbf{R}}_{t,a})$ denote the predicted pose and $(\mathbf{p}^{*}_{t,a},\mathbf{R}^{*}_{t,a})$ the reference pose obtained from the recorded action.
Translation and rotation errors are
\begin{align}
e^{\mathrm{trans}}_{t,a}
&=\left\lVert\hat{\mathbf{p}}_{t,a}-\mathbf{p}^{*}_{t,a}\right\rVert_2,\\
e^{\mathrm{rot}}_{t,a}
&=\arccos\!\left[
\operatorname{clip}\!\left(
\frac{\operatorname{tr}\!\left((\mathbf{R}^{*}_{t,a})^{\mathsf T}\hat{\mathbf{R}}_{t,a}\right)-1}{2},-1,1
\right)\right]\frac{180}{\pi}.
\end{align}
The units of $e^{\mathrm{trans}}_{t,a}$ and $e^{\mathrm{rot}}_{t,a}$ are mm and degrees, respectively.
For $T$ evaluated frames, table entries are arithmetic means:
\begin{equation}
E^{(\cdot)}_{a}=\frac{1}{T}\sum_{t=1}^{T}e^{(\cdot)}_{t,a}.
\end{equation}
After excluding first-frame JIT, the sequence contributes 375 samples.
Mean inference times are $61.571$ ms for standard 10-NFE and $21.923$ ms for two-stage 2-NFE, corresponding to a $2.808\times$ speedup and a $64.39\%$ reduction in time.

\begin{table}[!htbp]
    \centering
    \tablefont
    \caption{\textbf{Supplementary action-error metrics on the fixed validation sequence.}}
    \label{tab:app_two_stage_summary}
    \renewcommand{\arraystretch}{1.22}
    \begin{tabularx}{\linewidth}{@{}Lcc@{}}
        \toprule
        \textbf{Metric} & \makecell{\textbf{Standard $\pi_{0.5}$,}\\\textbf{10 NFE}} & \makecell{\textbf{Two-stage,}\\\textbf{2 NFE}} \\
        \midrule
        12-joint 95th-percentile absolute error (P95, rad) & 0.019209 & 0.018704 \\
        12-joint maximum absolute error (rad) & 0.057459 & 0.052449 \\
        Left/right gripper MAE (command unit) & 0.000482 / 0.000634 & 0.000548 / 0.000344 \\
        Left/right TCP rotation MAE ($^\circ$) & 1.047 / 0.902 & 0.911 / 1.169 \\
        \bottomrule
    \end{tabularx}
\end{table}

\FloatBarrier
\subsection{Per-Joint Error and Frame-Wise Variation}

\begin{figure}[H]
    \centering
    \includegraphics[width=\textwidth]{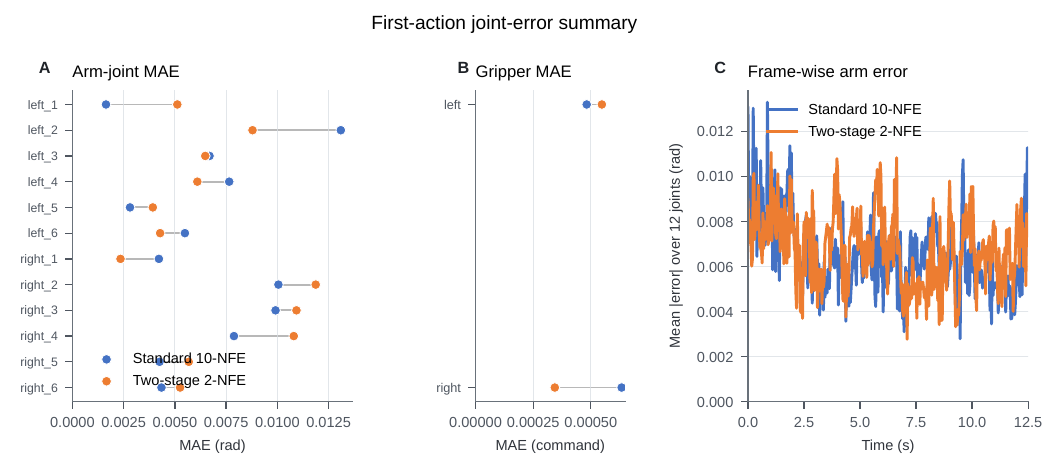}
    \caption{\textbf{Per-joint action errors on the fixed validation sequence.} (A) MAE for 12 rotational joints. (B) Left and right gripper-command MAE. (C) Frame-wise mean absolute error over the 12 rotational joints. Blue denotes the standard 10-NFE model, and orange denotes the two-stage 2-NFE model.}
    \label{fig:app_two_stage_joint_error}
\end{figure}

Figure~\ref{fig:app_two_stage_joint_error} reports per-joint MAE, left and right gripper MAE, and frame-wise mean absolute error over the 12 joints.
The two-stage model has P95 and maximum absolute errors of $0.018704$ rad and $0.052449$ rad, respectively, both lower than the standard model.
The panels show differences for individual joints and grippers.

\subsection{Fourteen-Dimensional Per-Joint Action Trajectories}

\begin{figure}[H]
    \centering
    \includegraphics[width=0.97\textwidth]{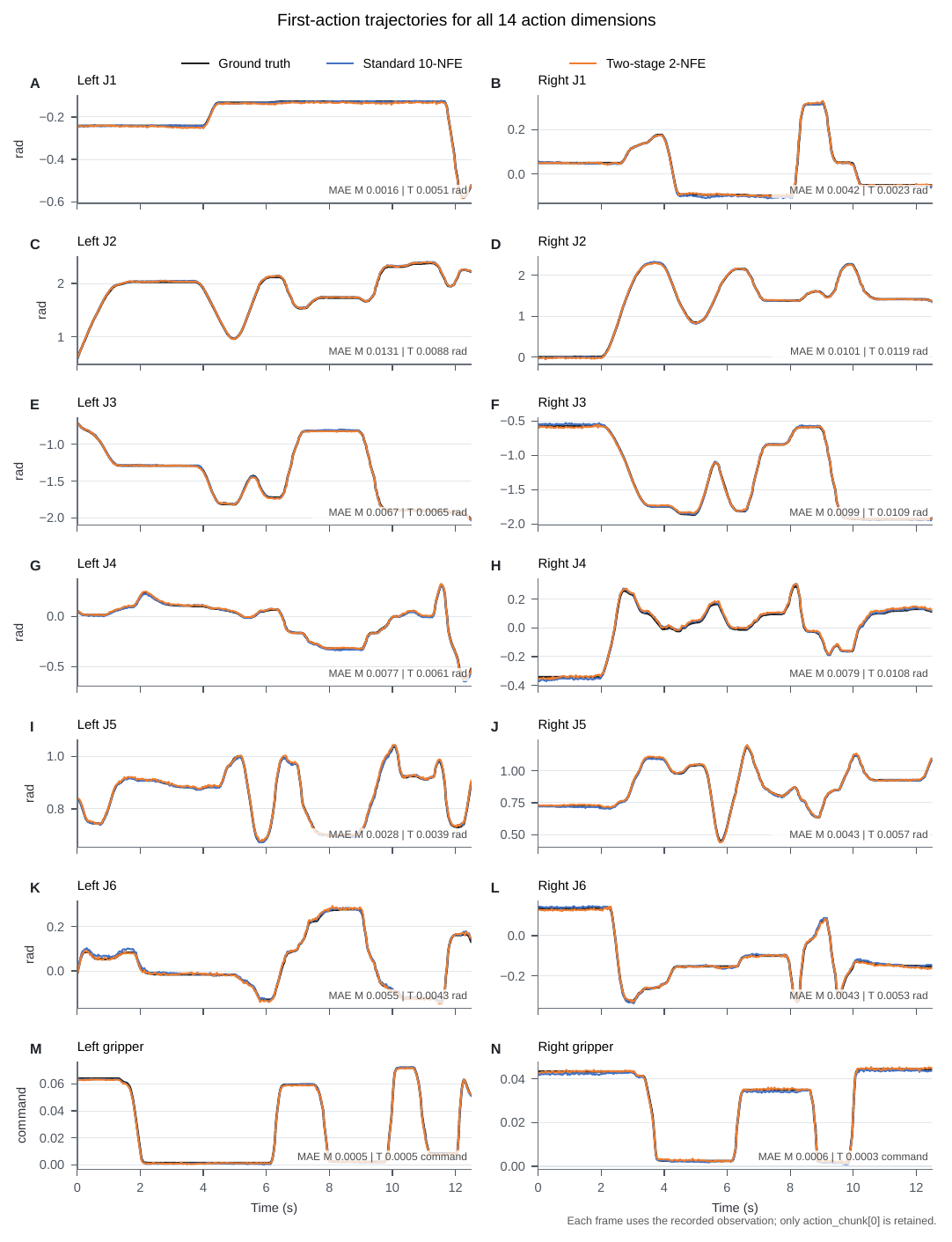}
    \caption{\textbf{Per-joint comparison.}}
    \label{fig:app_two_stage_joint_trajectories}
\end{figure}

Figure~\ref{fig:app_two_stage_joint_trajectories} shows 14-dimensional recorded and predicted action trajectories.
Black curves denote recorded actions; blue and orange curves denote predictions from the standard 10-NFE and two-stage 2-NFE models.
The left and right columns show six rotational joints and one gripper command for the respective arms.
M/T indicates the per-dimension MAE of the standard and two-stage models, respectively.
Across joints and grippers, the two-stage predictions broadly retain the main trends of the recorded actions.

\clearpage
\subsection{Bimanual TCP Trajectories and Pose Errors}

\begin{figure}[H]
    \centering
    \includegraphics[width=\textwidth]{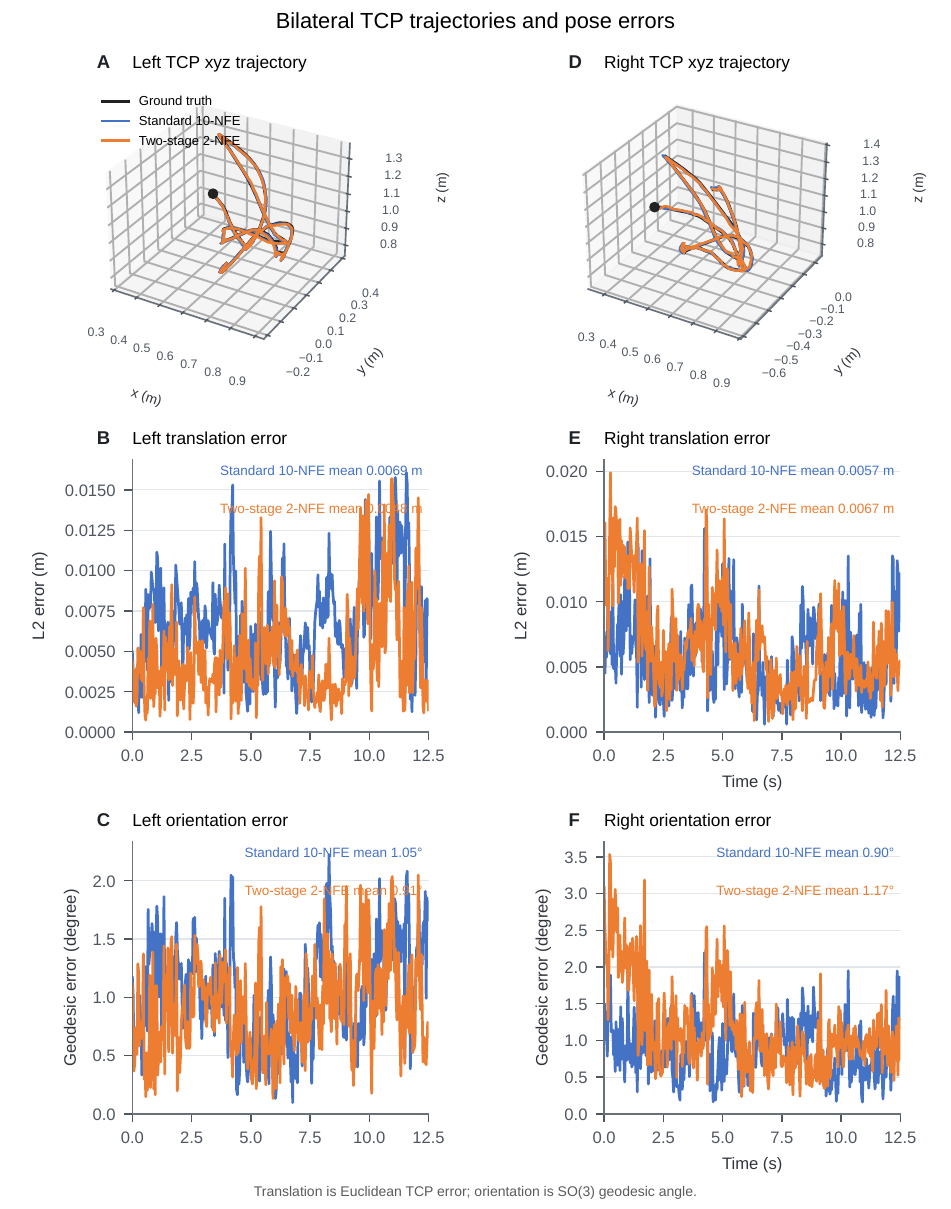}
    \caption{\textbf{Bimanual TCP trajectories and pose errors on the fixed validation sequence.} (A,D) Three-dimensional left and right TCP trajectories. (B,E) Euclidean translation error. (C,F) $\mathrm{SO}(3)$ geodesic orientation error. Black denotes the TCP trajectory obtained by applying URDF forward kinematics to the recorded action. Blue and orange denote the standard 10-NFE and two-stage 2-NFE models.}
    \label{fig:app_two_stage_tcp_error}
\end{figure}

As shown in \figref{fig:app_two_stage_tcp_error}, two-stage denoising reduces left-arm TCP rotation MAE from $1.047^\circ$ to $0.911^\circ$, while increasing the right-arm value from $0.902^\circ$ to $1.169^\circ$.
The three-dimensional trajectories and frame-wise translation errors further show temporal differences between the models.

%% file: sections/appendix_e_execution_en.tex
\section{Additional Execution Figures}
\label{app:execution_figures}

\newcommand{\executionpanelwidth}{0.45\linewidth}
\newcommand{\executionpanelgap}{\hspace{0.04\linewidth}}

\begingroup
\setlength{\parskip}{0.18em}
\setlength{\intextsep}{9pt plus 2pt minus 2pt}
\captionsetup{skip=4pt}
\begin{figure}[H]
    \centering
    \begin{subfigure}[t]{\executionpanelwidth}
        \vspace{0pt}\centering
        \includegraphics[width=\linewidth]{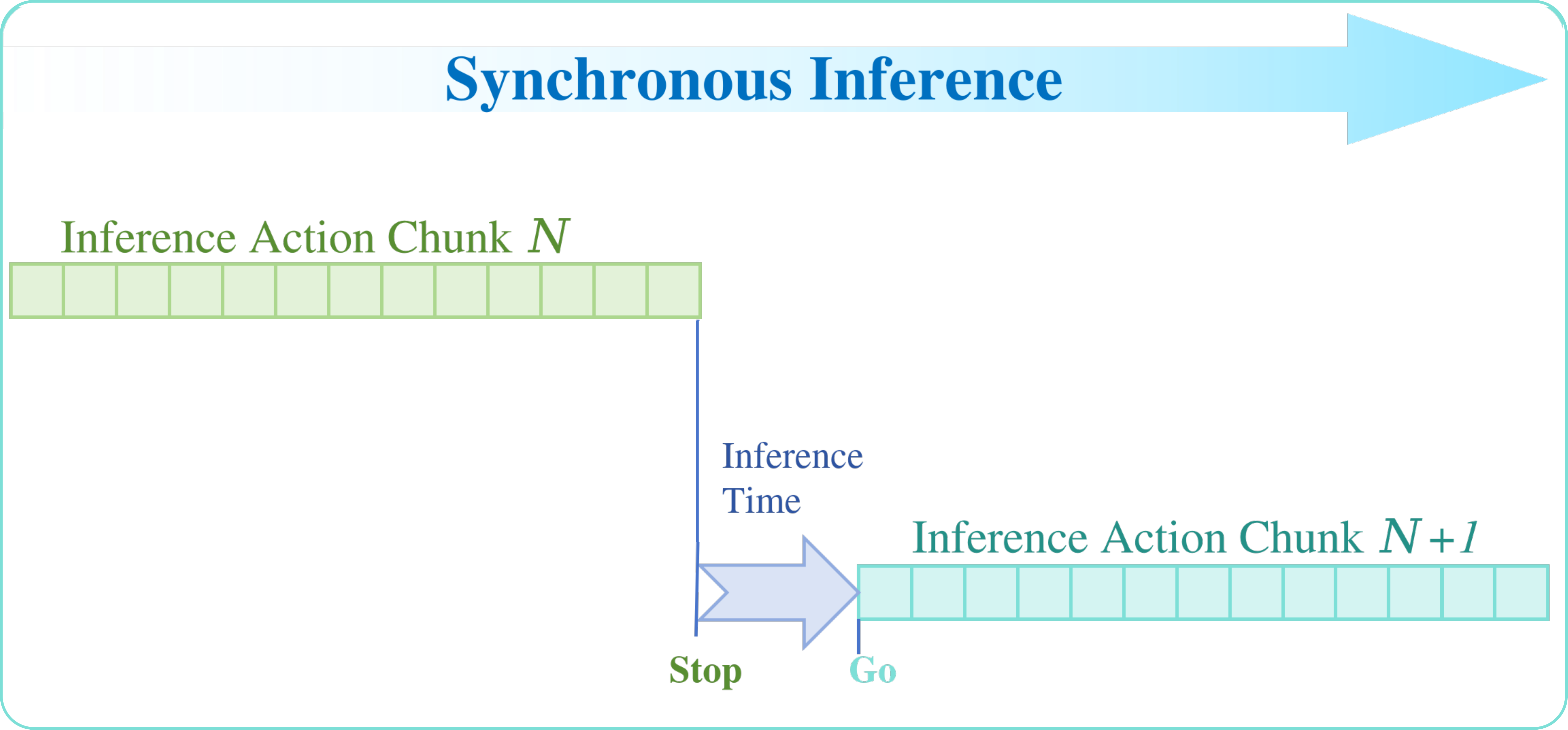}
        \caption{\textbf{Synchronous inference.}}
    \end{subfigure}\executionpanelgap%
    \begin{subfigure}[t]{\executionpanelwidth}
        \vspace{0pt}\centering
        \includegraphics[trim=0bp 0bp 6.76370bp 0bp,clip,width=\linewidth]{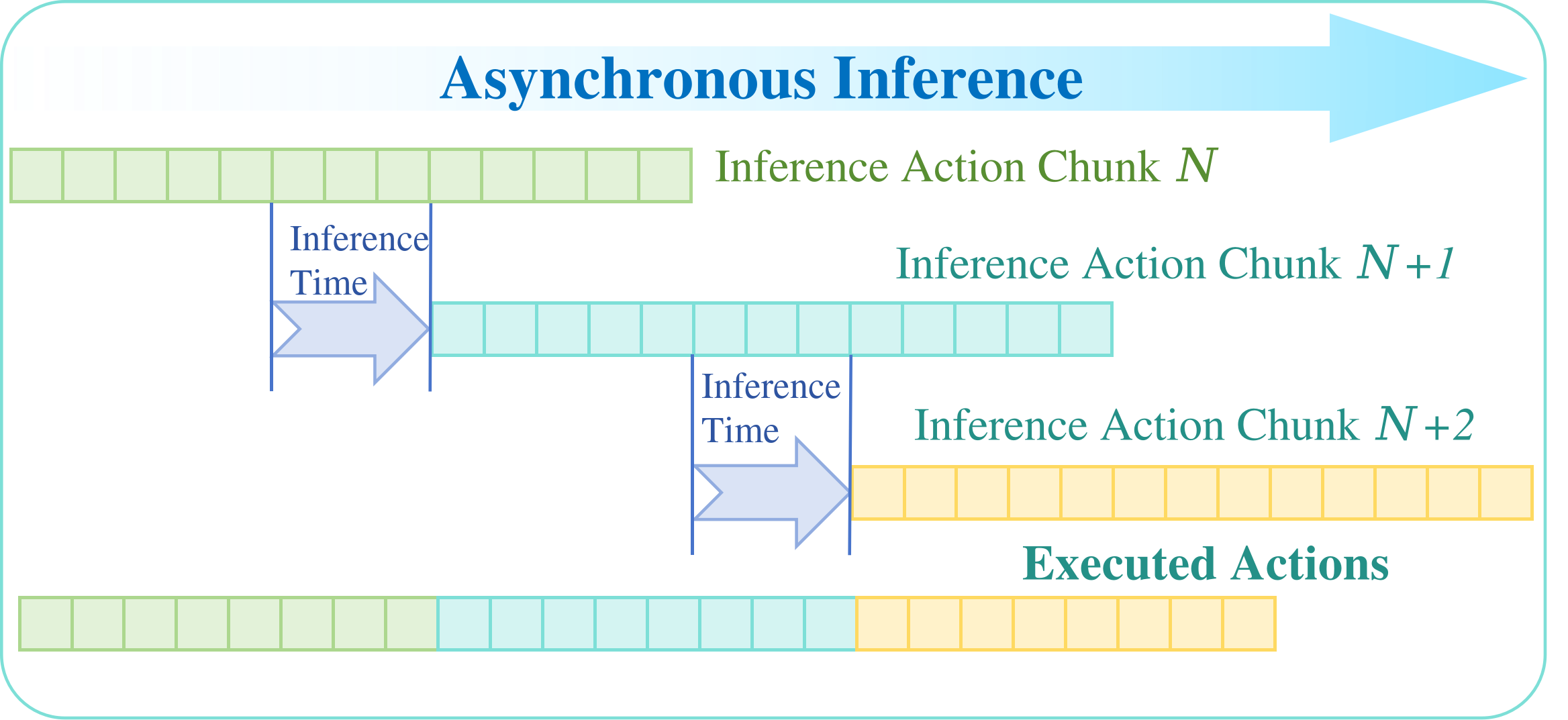}
        \caption{\textbf{Asynchronous inference.}}
    \end{subfigure}\par\medskip
    \begin{subfigure}[t]{\executionpanelwidth}
        \vspace{0pt}\centering
        \includegraphics[trim=0bp 0.21818bp 7.63644bp 0bp,clip,width=\linewidth]{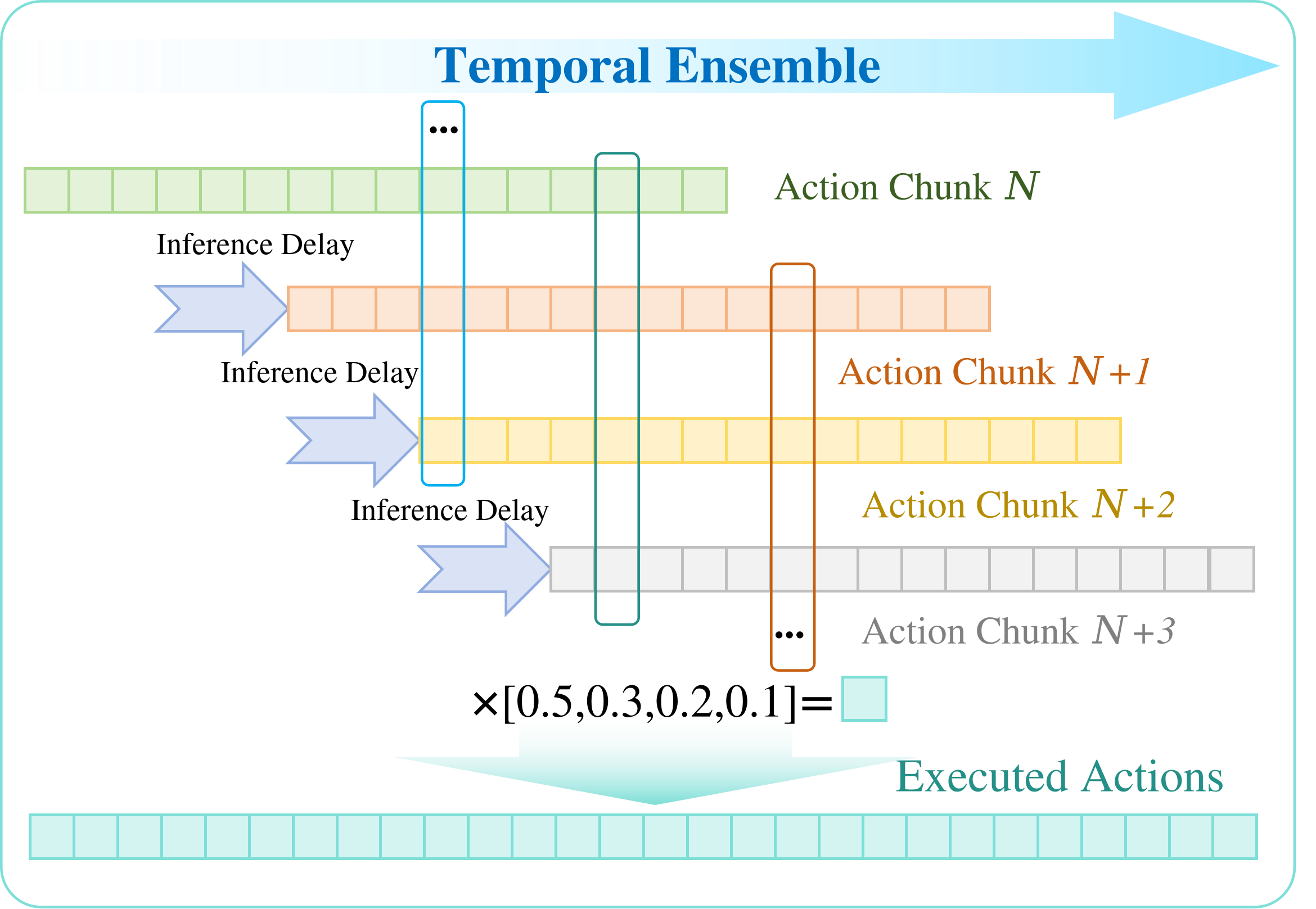}
        \caption{\textbf{Temporal Ensembling.}}
    \end{subfigure}\executionpanelgap%
    \begin{subfigure}[t]{\executionpanelwidth}
        \vspace{0pt}\centering
        \includegraphics[width=\linewidth]{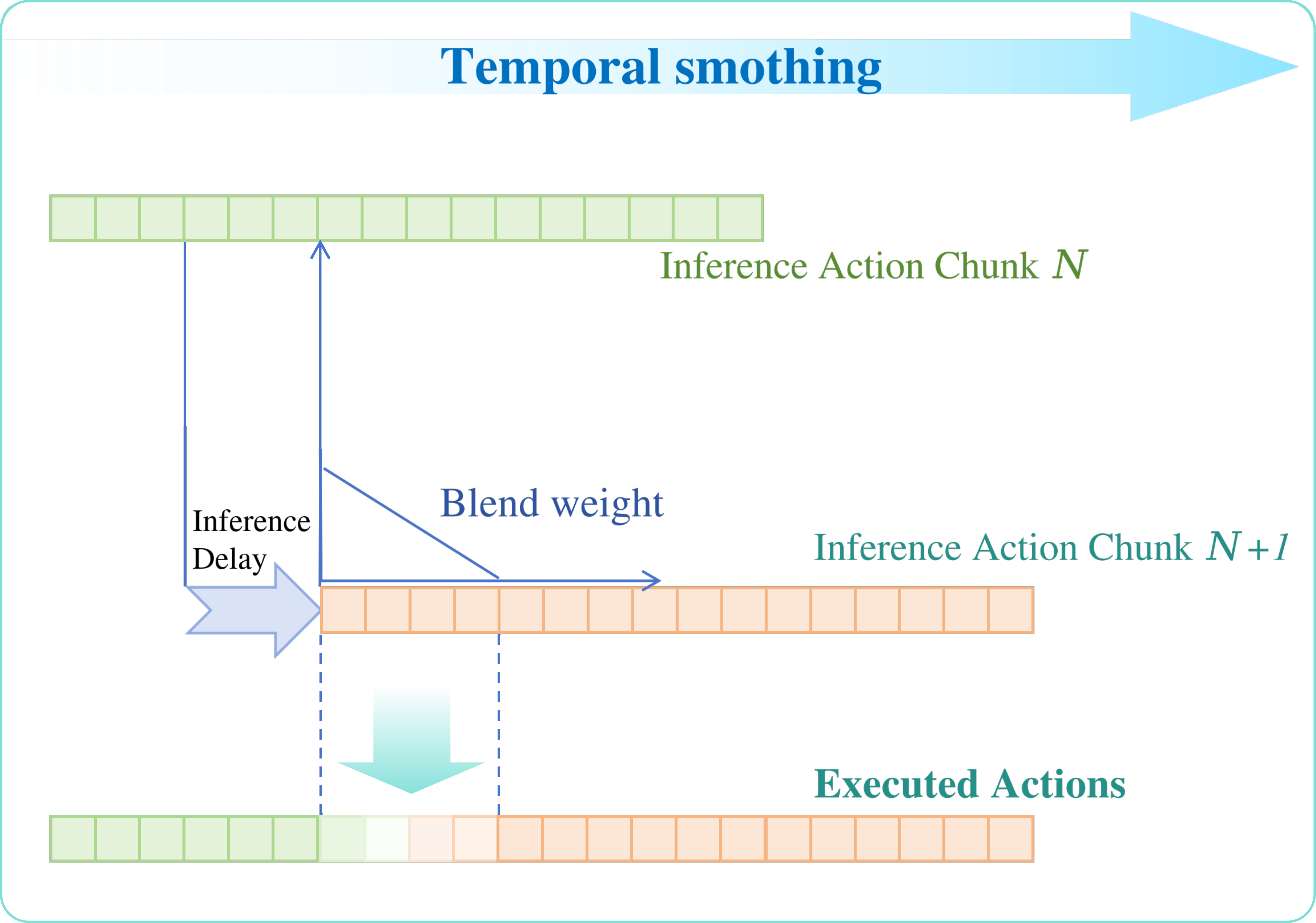}
        \caption{\textbf{Temporal Smoothing.}}
    \end{subfigure}
    \caption{\textbf{Reference inference and execution modes.}}
    \label{fig:inference_execution_modes}
\end{figure}

\begin{figure}[H]
    \centering
    \begin{subfigure}[t]{\executionpanelwidth}
        \vspace{0pt}\centering
        \includegraphics[width=\linewidth]{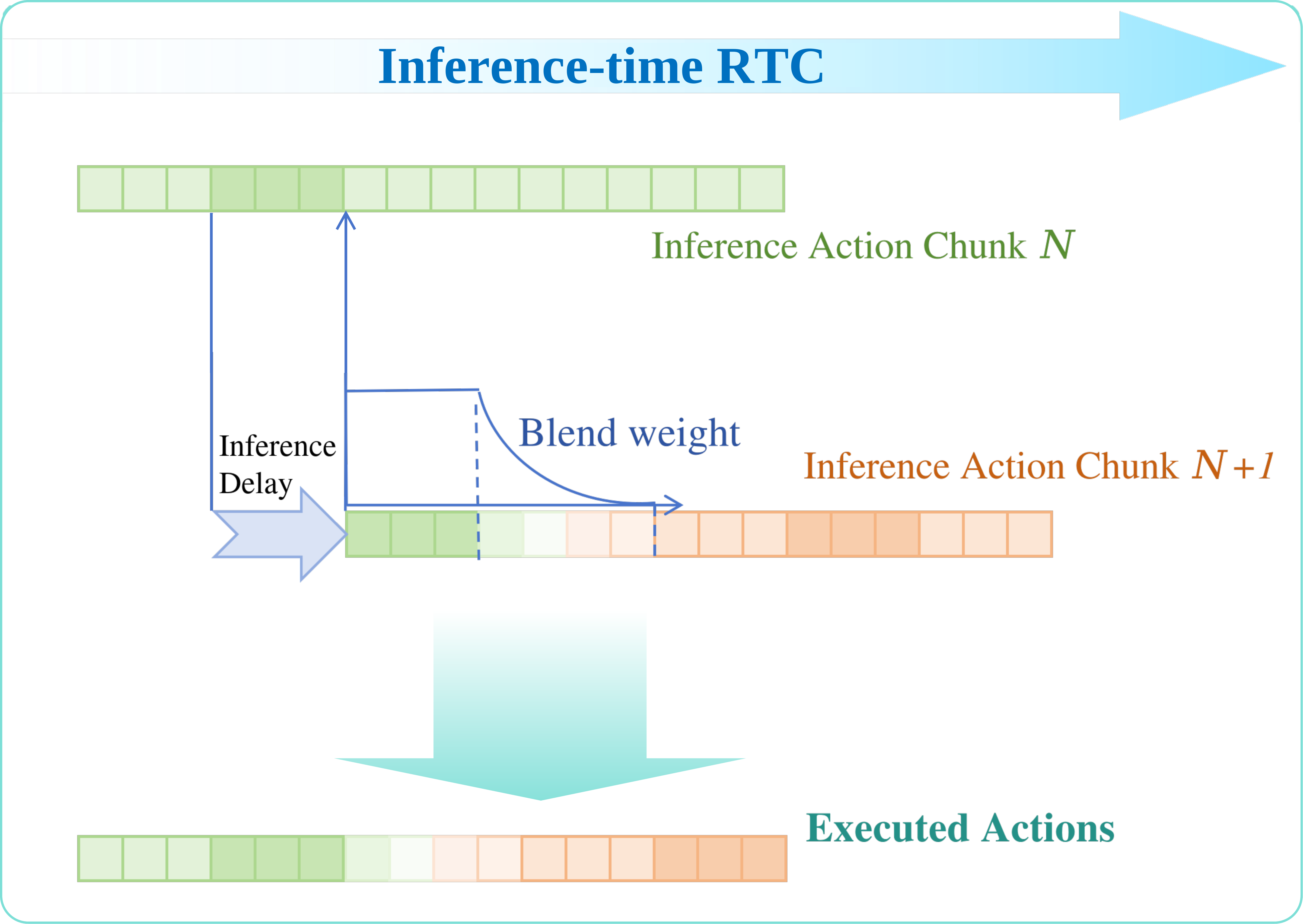}
        \caption{\textbf{Inference-time RTC.}}
    \end{subfigure}\executionpanelgap%
    \begin{subfigure}[t]{\executionpanelwidth}
        \vspace{0pt}\centering
        \includegraphics[trim=0bp 0bp 0.87274bp 0bp,clip,width=\linewidth]{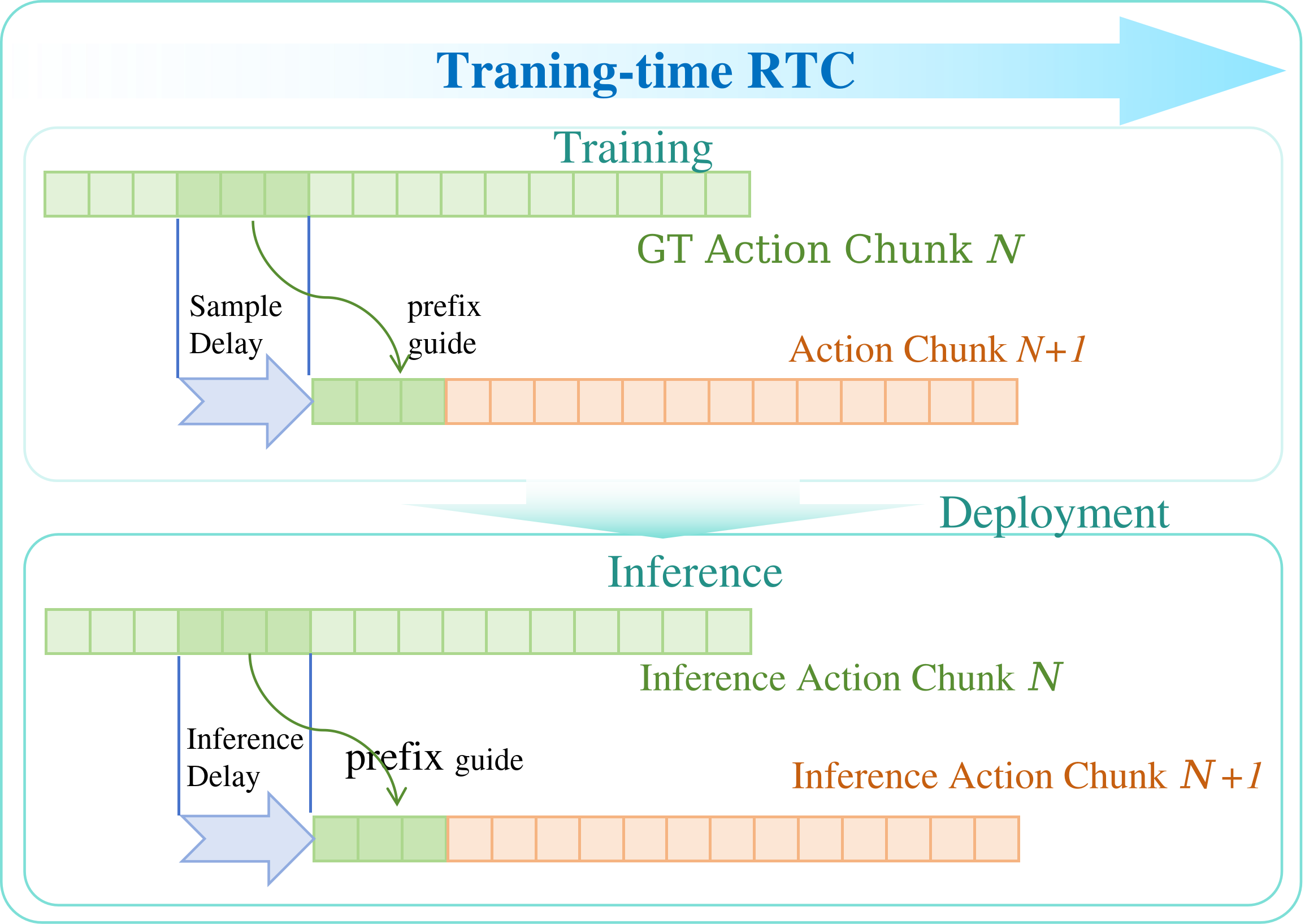}
        \caption{\textbf{Training-time RTC.}}
    \end{subfigure}\par\medskip
    \begin{subfigure}[t]{\executionpanelwidth}
        \vspace{0pt}\centering
        \includegraphics[trim=0bp 0bp 4.58186bp 0bp,clip,width=\linewidth]{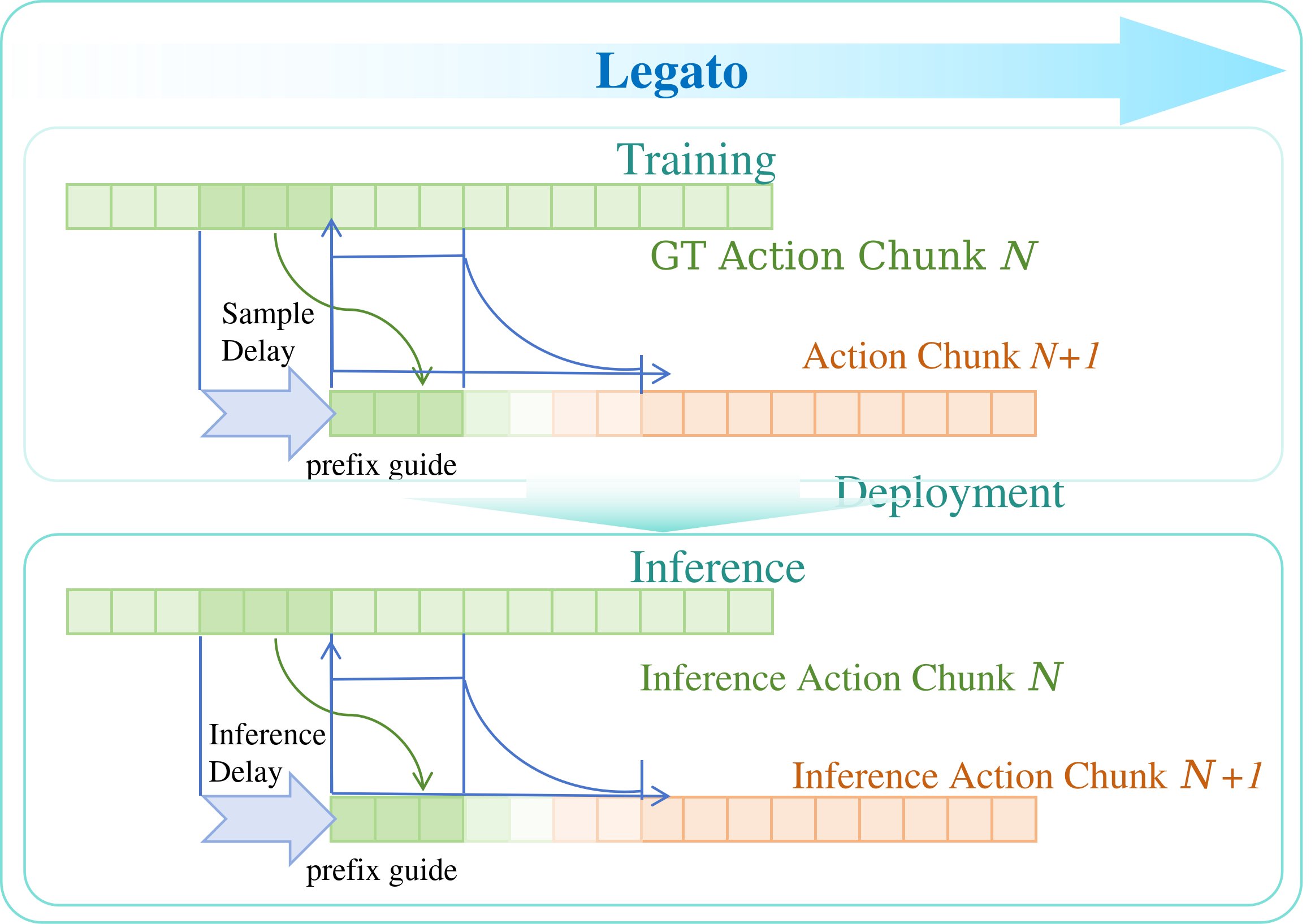}
        \caption{\textbf{Legato.}}
    \end{subfigure}\executionpanelgap%
    \begin{subfigure}[t]{\executionpanelwidth}
        \vspace{0pt}\centering
        \includegraphics[trim=0bp 0bp 21bp 0bp,clip,width=\linewidth]{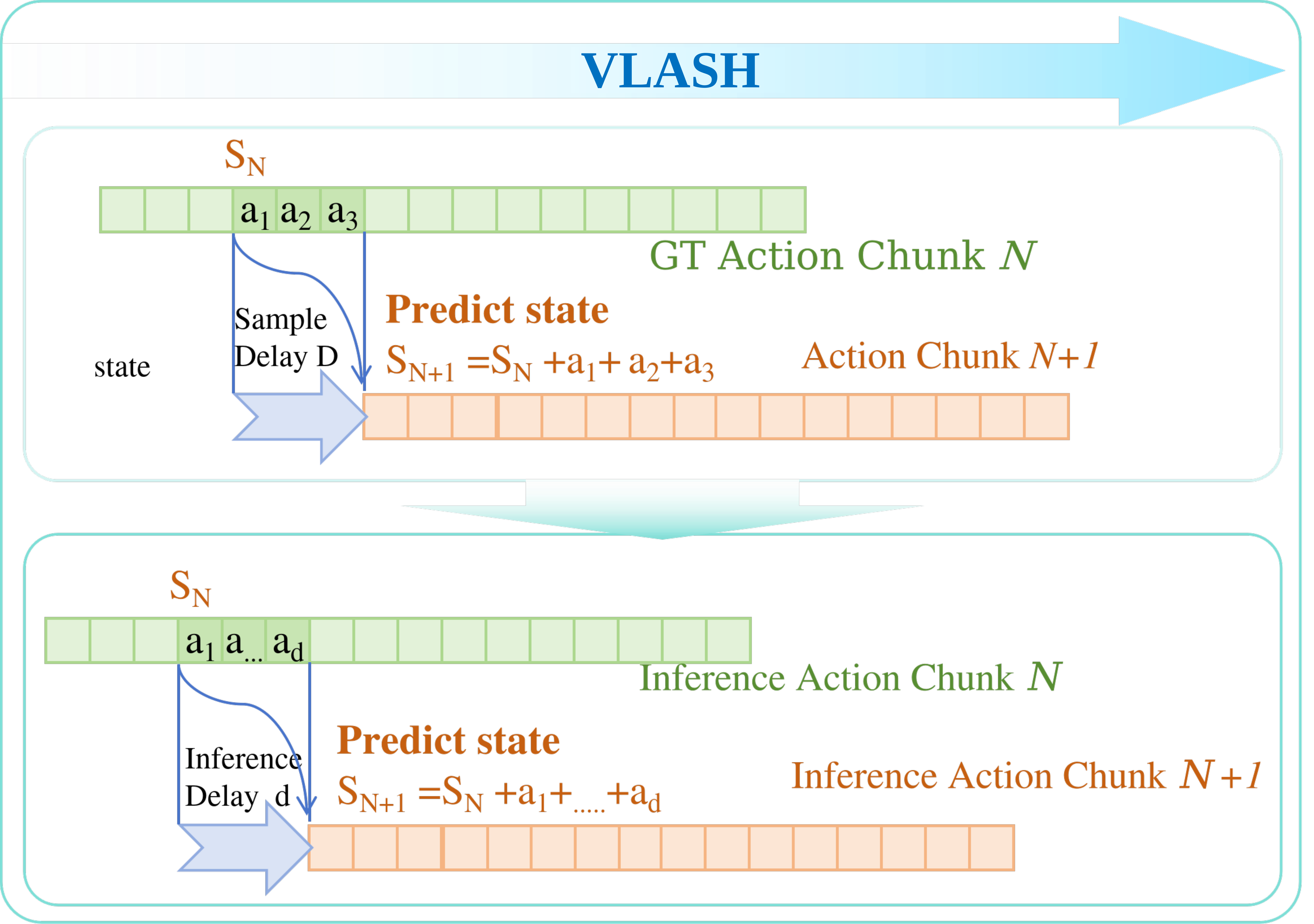}
        \caption{\textbf{VLASH.}}
    \end{subfigure}
    \caption{\textbf{Representative delay-aware execution methods.}}
    \label{fig:delay_aware_methods}
\end{figure}
\endgroup

\begin{figure}[H]
    \centering
    \includegraphics[width=\textwidth]{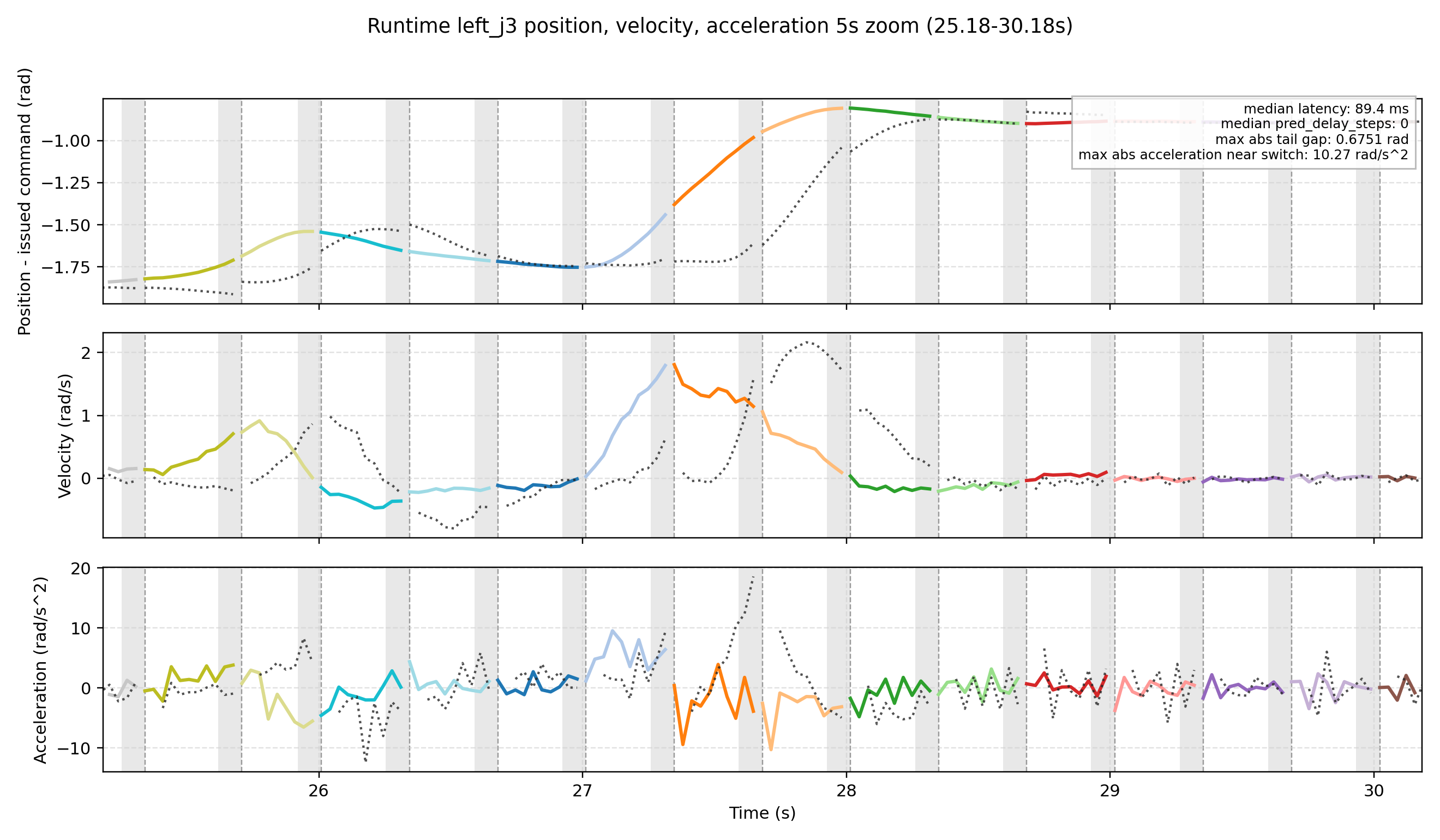}
    \caption{\textbf{Local \texttt{left\_j3} trajectory for Temporal Smoothing.}}
    \label{fig:local_smoothing}
\end{figure}

\begin{figure}[H]
    \centering
    \includegraphics[width=\textwidth]{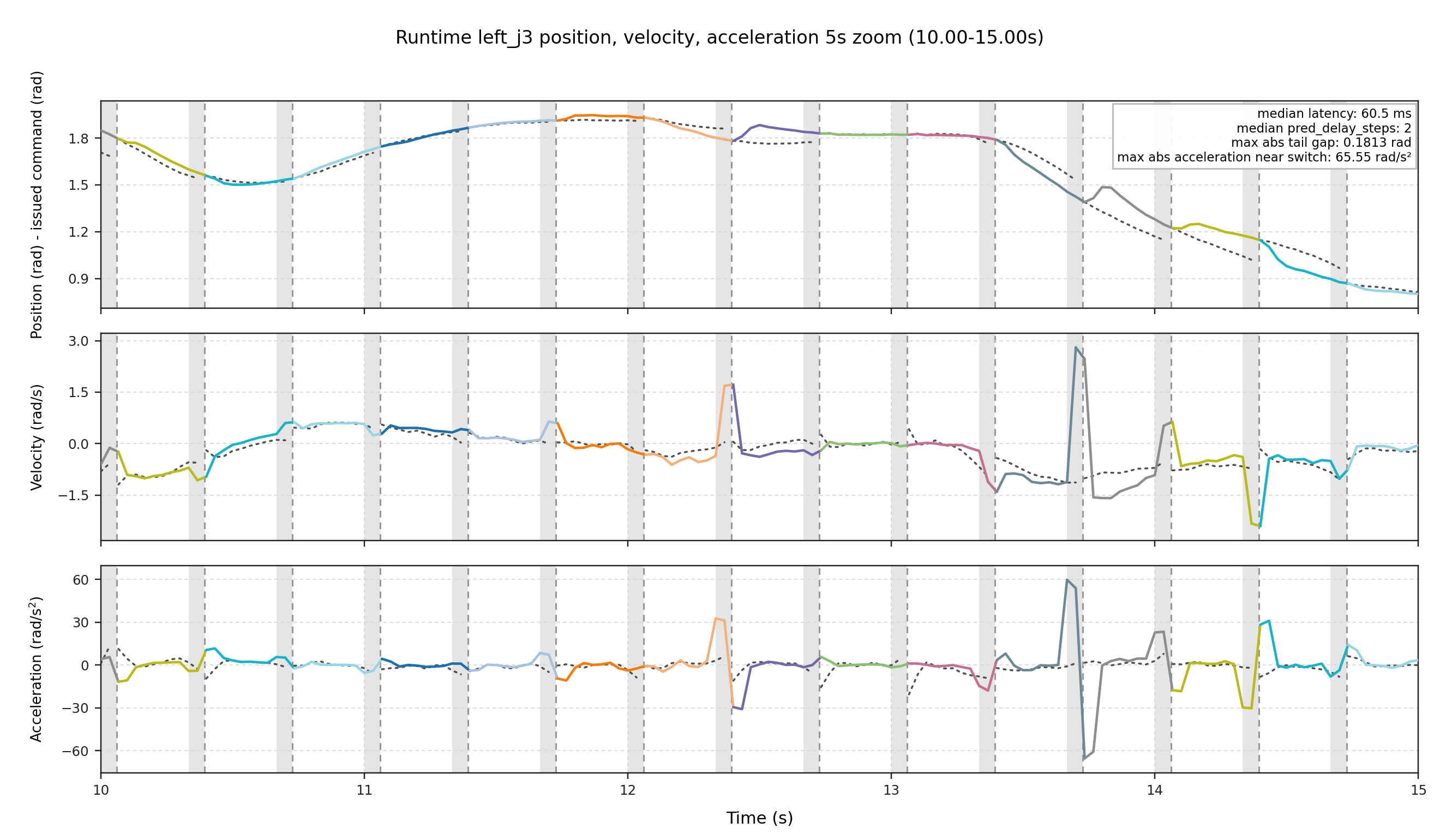}
    \caption{\textbf{Local \texttt{left\_j3} trajectory for Inference-time RTC.}}
    \label{fig:local_inference_rtc}
\end{figure}

\begin{figure}[H]
    \centering
    \includegraphics[width=\textwidth]{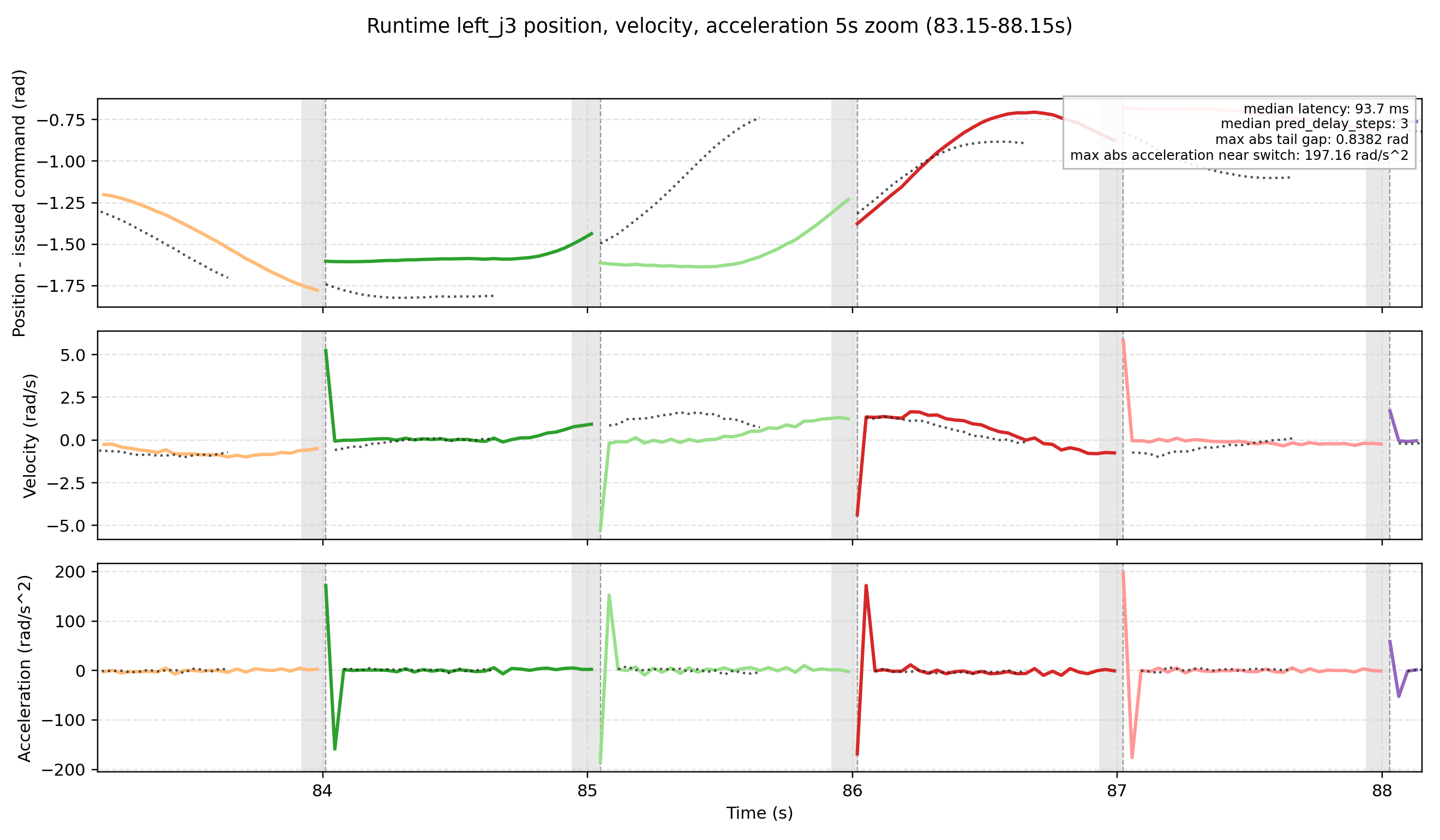}
    \caption{\textbf{Local \texttt{left\_j3} trajectory for Training-time RTC.}}
    \label{fig:local_training_rtc}
\end{figure}

\begin{figure}[H]
    \centering
    \includegraphics[width=\textwidth]{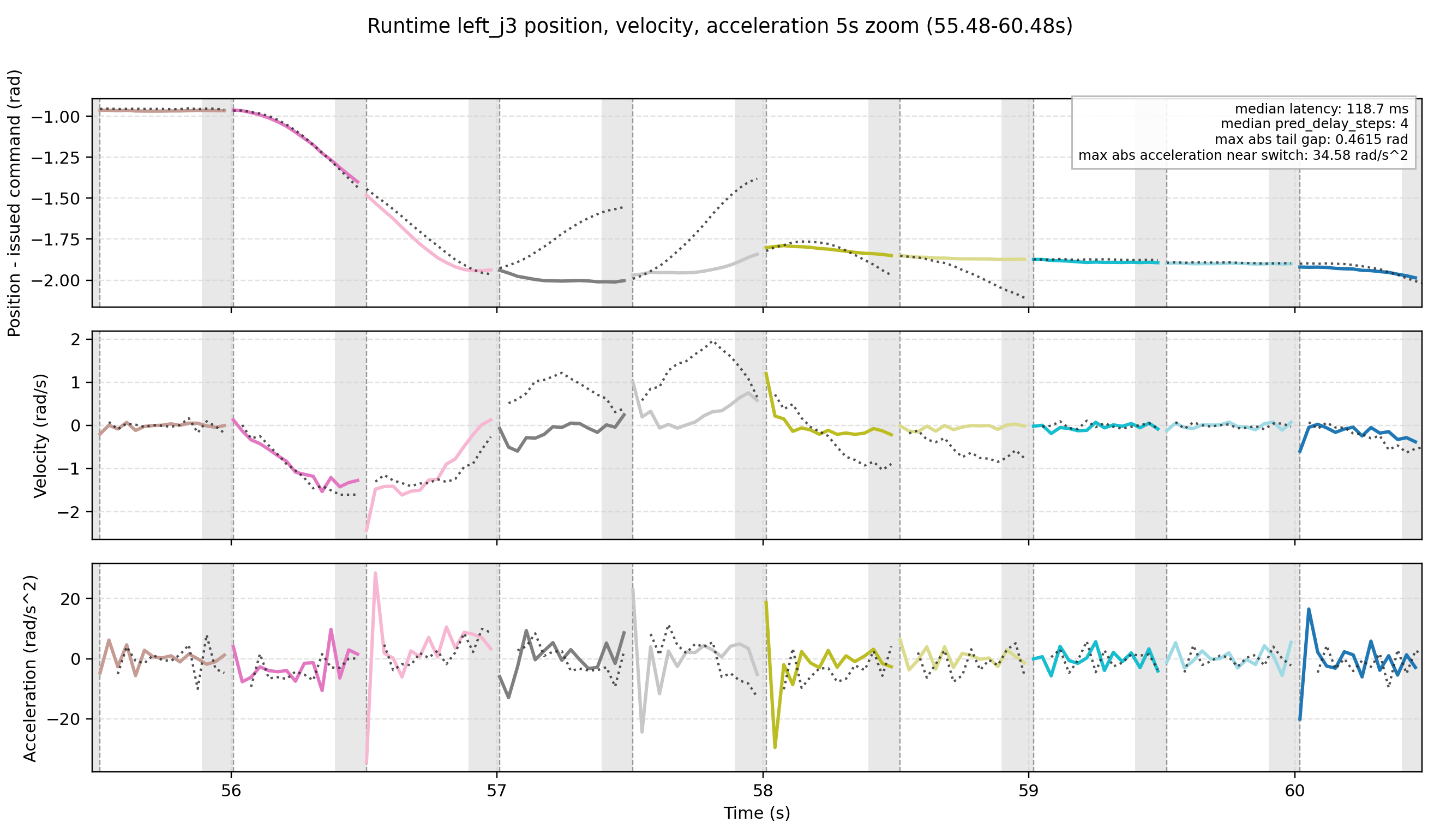}
    \caption{\textbf{Local \texttt{left\_j3} trajectory for VLASH.}}
    \label{fig:local_vlash}
\end{figure}